\pdfoutput=1

\documentclass[11pt]{article}

\usepackage{acl}
\usepackage{booktabs} 
\usepackage{multirow}
\usepackage{times}
\usepackage{latexsym}
\usepackage{amsmath}
\usepackage{amssymb}
\usepackage{amsfonts} 
\usepackage{textcomp}
\usepackage{todonotes}
\usepackage{subcaption}
\usepackage{svg}
\usepackage[T1]{fontenc}

\usepackage[utf8]{inputenc}
\usepackage{graphicx}
\usepackage{microtype}
\usepackage{soul}
\usepackage{pifont}
\title{Turn Waste into Worth: Rectifying Top-$k$ Router of MoE}

\author{Zhiyuan Zeng\textsuperscript{1,2\footnote{*}},\quad Qipeng Guo\textsuperscript{2},\quad Zhaoye Fei\textsuperscript{1,2},\quad Zhangyue Yin\textsuperscript{1},\quad Yunhua Zhou\textsuperscript{2} \\
    {\bf Linyang Li\textsuperscript{2},\quad Tianxiang Sun\textsuperscript{1},\quad Hang Yan\textsuperscript{2\ding{169}},\quad Dahua Lin\textsuperscript{2},\quad Xipeng Qiu\textsuperscript{1\ding{169}}} \\
$^1$School of Computer Science, Fudan University, Shanghai, China \\
$^2$Shanghai AI Laboratory \\
  \texttt{\{cengzy23,yinzy21\}@m.fudan.edu.cn} \\
  \texttt{\{zyfei20, zhouyh20, txsun19, xpqiu\}@fudan.edu.cn} \\
  \texttt{\{guoqipeng, lilinyang, yanhang\}@pjlab.org.cn}
  }

\begin{document}
\maketitle
\def\thefootnote{*}\footnotetext{Work done during internship at Shanghai AI Laboratory.}
\def\thefootnote{\ding{169}}\footnotetext{Corresponding Authors.}

\begin{abstract}
Sparse Mixture of Experts (MoE) models are popular for training large language models due to their computational efficiency. However, the commonly used top-$k$ routing mechanism suffers from redundancy computation and memory costs due to the unbalanced routing. Some experts are overflow, where the exceeding tokens are dropped. While some experts are vacant, which are padded with zeros, negatively impacting model performance. To address the dropped tokens and padding, we propose the Rectify-Router, comprising the Intra-GPU Rectification and the Fill-in Rectification. The Intra-GPU Rectification handles dropped tokens, efficiently routing them to experts within the GPU where they are located to avoid inter-GPU communication. The Fill-in Rectification addresses padding by replacing padding tokens with the tokens that have high routing scores. Our experimental results demonstrate that the Intra-GPU Rectification and the Fill-in Rectification effectively handle dropped tokens and padding, respectively. Furthermore, the combination of them achieves superior performance, surpassing the accuracy of the vanilla top-1 router by 4.7\%. 
\end{abstract}

\section{Introduction}
Sparse Mixture of Experts (MoE) is gaining popularity as a model architecture for training large language models~\cite{switchc,glam,st-moe,mixtral,deepseek-moe} owing to its computational efficiency. In a sparse MoE model, each token is assigned to one or more experts based on a routing mechanism. The top-$k$ router is currently the most widely used routing mechanism, where tokens are directed to the experts with the top-$k$ scores.
 
However, top-$k$ router is unbalanced, where the number of tokens routed to different GPUs is not the same. In order to achieve a balanced workload across GPUs, top-$k$ routing imposes a maximum limit on the number of tokens that each expert can process. Consequently, any tokens exceeding this limit are dropped, and vacant experts are padded with zeros, which negatively impacts the overall model performance~\citep{megablock}. 

\begin{figure}
    \centering
    \includegraphics[width=0.5\textwidth]{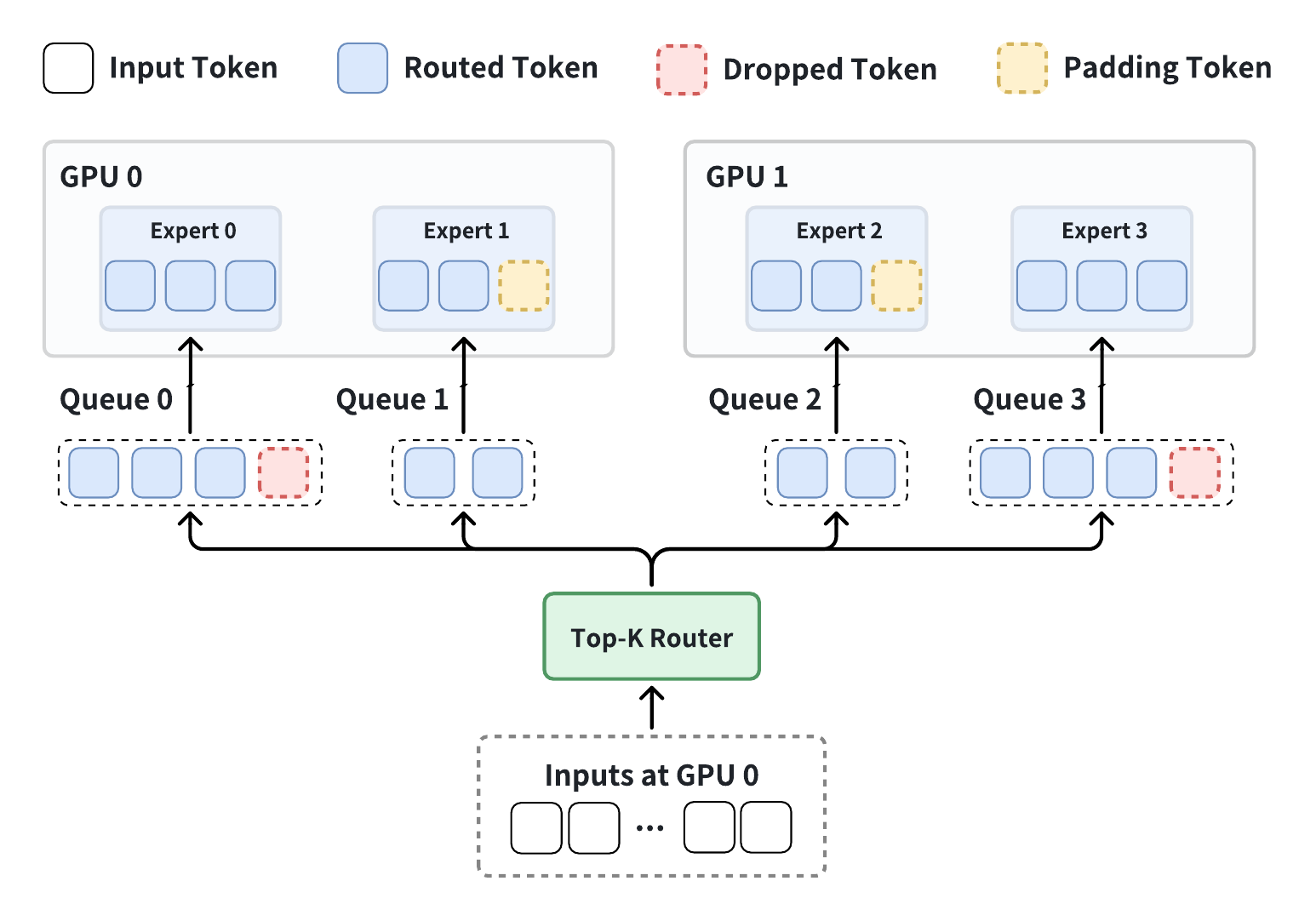}
    \caption{The illustration of dropped token and padding in top-$k$ router of MoE. Queue $i$ represents the queue of tokens to be sent to expert $i$. The capacity of each expert is fixed to 3. }
    \label{fig:problem}
\end{figure}

Previous studies have attempted to address the balance issue in routing by introducing auxiliary loss mechanisms~\cite{first-moe,gshard,st-moe}. But even with these enhancements, the performance drop resulting from dropped tokens is still significant~\citep{expert-choices,megablock}. Although some approaches have proposed absolutely balanced routers, they have been found to underperform the original top-$k$ routing methodology~\cite{mlp-moe}.

Rather than focusing on improving the balance of the top-$k$ router, this work introduces an alternative approach called the \textbf{Rectify-Router}, which rectifies top-$k$ router by post-processing the dropped tokens and padding from the top-$k$ router. We propose two Rectify-Routers: the \textbf{Intra-GPU Rectification} and the \textbf{Fill-in Rectification}. The Intra-GPU Rectification is designed to handle the dropped tokens, while the Fill-in Rectification specifically addresses the padding issue. 

Post-processing the dropped tokens with another router may bring expensive communication cost. Therefore, we propose the Intra-GPU Rectification which routes the dropped tokens to the experts within the GPU where they are located, eliminating the need for inter-GPU communication. Our empirical experiments have demonstrated that the Intra-GPU Rectification effectively handles the post-processing of dropped tokens and is more efficient than the commonly used routers, in terms of communication.

To address the padding issue, we present the Fill-in Rectification, which replace padding tokens with the tokens that have high routing scores. Fill-in Rectification first identifies the optimal expert for each token based on the routing scores and subsequently selects the tokens with the highest routing score to replace the padding for each expert. By employing Fill-in Rectification, tokens with the higher routing scores receive more computational allocation.

The Intra-GPU Rectification and Fill-in Rectification are orthogonal approaches that can be seamlessly combined. Our experiments have demonstrated their effectiveness in handling dropped tokens and padding. Furthermore, combing the Intra-GPU Rectification and Fill-in Rectification yield improved performance compared to using them individually. 

\paragraph*{Contributions} The contributions of our work can be summarized as follows:
\begin{enumerate}
    \item We introduce the concept of Rectify-Router to handle the dropped tokens and padding in MoE models. Specifically, the dropped tokens are efficiently processed using the Intra-GPU Rectification, while the padding tokens are optimally managed using the Fill-in Rectification.
    \item Our experiments demonstrate that both the Intra-GPU Rectification and the Fill-in Rectification significantly improve the performance of the top-$k$ routing, even without additional training.
    \item We found that our methods are robust to various settings of expert capacity and that Intra-GPU Rectification can be used for accelerating MoE by reducing expert capacities.
\end{enumerate}

\section{Related Works}
The routing of MoE can be classified into two categories: balanced and unbalanced. The balanced routing assigns the same number of tokens to each expert, while the unbalanced routing does not make sure that the number of tokens received by each expert is the same. 

\begin{figure*}[t]
    \centering 
    \includegraphics[width=1.0\textwidth]{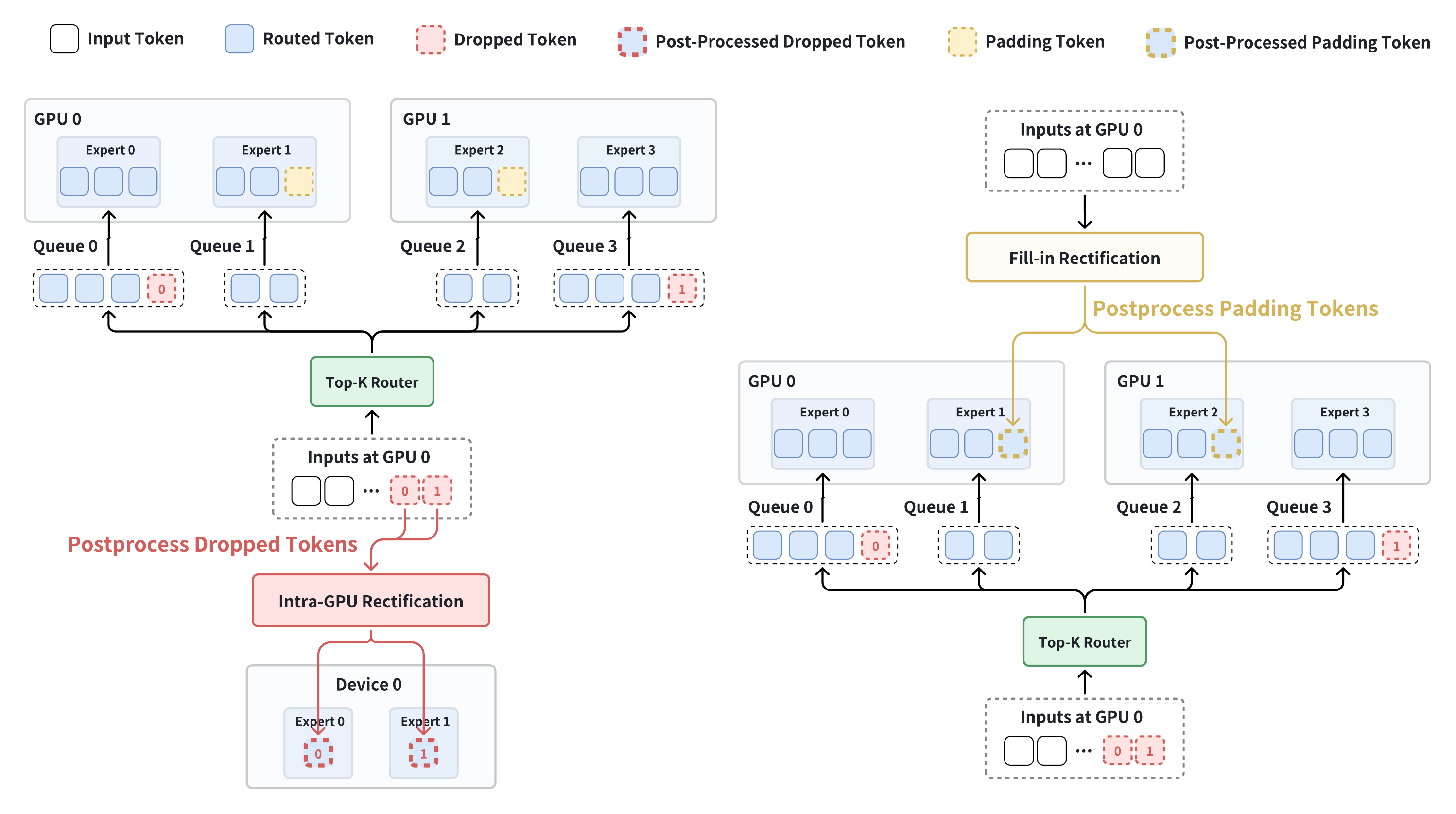}
    \caption{Left: Post-processing of dropped tokens at GPU 0 with Intra-GPU Rectification. Right: Post-processing of padding at GPU 0 with Fill-in Rectification. }
    \label{fig:solution}
\end{figure*}

\paragraph{Unbalanced Routing} Top-$k$ routing was the most commonly used unbalanced routing proposed by~\citet{first-moe}, which greedily assigns tokens to experts, according to the token-expert assignment scores. Numerous MoE models have adopted top-$k$ routing, including Switch Transformer~\cite{switchc}, Glam~\cite{glam}, ST-MoE~\cite{st-moe}, Flan-MoE~\cite{flan-moe}, and NLLB~\cite{nllb}, to name just a few. 

It is worth noting that many unbalanced routing methods are variations or derivatives of top-$k$ routing. For example, Switch Transformer~\cite{switchc} argues in favor of using top-1 routing instead of top-2 routing for improved efficiency. ST-MoE~\cite{st-moe} and LIMoE~\cite{mmoe} propose auxiliary loss functions to enhance the stability of MoE during training. Additionally, SCoMoE~\cite{scomoe} and Gating-Dropout~\cite{moe-dropout} improve the efficiency of top-$k$ routing by designing hierarchical routing systems based on the hierarchical structure of the communication topology.

The routing method proposed in this paper is also a variation of top-$k$ routing. However, unlike the aforementioned approaches, our objective is to address the issues of dropped tokens and padding that arise from unbalanced routing. Switch Transformer~\citep{switchc} tackles the problem of dropped tokens by increasing the capacity of experts, allowing each expert to handle more tokens. While this approach reduces the number of dropped tokens, it introduces additional overhead in terms of both speed and memory. On the other hand, Megablocks~\cite{megablock} addresses the challenges of padding and dropped tokens by gathering all experts onto the same GPU and employing model parallelism rather than expert parallelism. However, their approach may encounter difficulties when dealing with a large expert size or a substantial number of experts.

\paragraph{Balanced Routing} In response to the imbalance issue inherent in top-$k$ routing, several balanced routing methods have been proposed. For instance, the Base Layer approach~\cite{base-layer} employs a balanced assignment algorithm to evenly distribute tokens among experts. However, their assumption that tokens within the same batch can be evenly clustered may not hold true in all cases, which can potentially result in poorer performance~\cite{mlp-moe}. Another alternative to balanced routing is random routing~\cite{random-moe}, which assigns tokens to experts in a random manner. While random routing achieves balance and efficiency, it lacks any specialization or optimization in the routing process. Another approach called expert choices~\cite{expert-choices} allows each expert to select a fixed number of tokens, rather than relying on tokens to determine their target experts. This approach helps to avoid padding issues but still results in dropped tokens. Soft routing~\cite{soft-moe} is a method that compresses tokens by applying a linear transformation to generate fixed-size hidden states for each expert. However, this method is only suitable for encoder models with fixed input lengths and may not be applicable to autoregressive decoder models. 

\section{Preliminary}
In this section, we will introduce expert parallelism, top-$k$ routing, and two prevalent challenges that emerge while employing top-$k$ routing: padding and dropped tokens.

\paragraph{Expert Parallelism and Top-$k$ Routing} In expert parallelism, experts are distributed across GPUs uniformly. If there are $n$ experts and $k$ GPUs, each GPU contains $k/n$ experts. The process of transmitting tokens to the respective experts entails inter-GPU communication.

Top-$k$ routing greedily assigns tokens to experts according to the routing score:
\begin{equation}
\mathbb{R}_i = \text{argtopk}_{j \in [m]} \{ a_{ij} | a_{ij} = w_j^T x_i \}
    \label{eq:route-score}
\end{equation}
where $a_{ij}$ is the score of assigning the $i$th token to the $j$th expert, $w_j$ denotes the embedding vector of the $j$th expert, $x_i$ corresponds to the hidden states of the $i$ token. The index set  $\mathbb{R}_i$ signifies the target experts of the $i$th token. Given the scores of assigning token $x_i$ to $m$ experts, denoted as $a_{i0}, a_{i1}, ..., a_{im}$, $\mathbb{R}_i$ contains the indices of experts with top-$k$ scores.

Since each token undergoes processing by multiple experts, the outputs of these experts for the same token are consolidated through linear combination. The combining weights are determined by the normalized routing scores, as defined in Eq.~\eqref{eq:route-score}:
\begin{equation}
    o_i=\sum_{j\in \mathbb{R}_i} {\frac{e^{a_{ij}}}{\sum_{j\in\mathbb{R}_i} e^{a_{ij}}} {E_j(x_i)}}.
    \label{eq:norm}
\end{equation}
Here, $o_i$ represents the combined result of token $x_i$. The term $\frac{e^{a_{ij}}}{\sum_j^k e^{a_{ij}}}$ denotes the normalized routing scores, while $E_j(x_i)$ refers to the outputs of the $j$th expert with token $x_i$ as its input.

The top-$k$ routing approach exhibits an inherent imbalance, wherein the distribution of tokens among different experts is not uniform. However, the current distributed framework exclusively supports balanced computation across GPUs. Consequently, there exists a limitation on the maximum number of tokens that each expert can receive, which is referred to as the capacity. The capacity is determined by the capacity factor, which is typically set to $k$ for top-$k$ routing~\cite{gshard,deepspeed-moe}. Mathematically, the capacity can be expressed as:
$$
\text{capacity}=\text{capacity factor} \times \frac{\text{number of tokens}}{\text{number of experts}}.
\label{capacity}
$$

\paragraph{Dropped Tokens and Padding}
The issue of dropped tokens and padding arises naturally when dealing with the expert capacity setting, as depicted in Figure~\ref{fig:problem}. With a fixed expert capacity, overflow experts are compelled to drop tokens with the lowest routing scores and directly pass them to the next layer through residual connections, as highlighted in red in Figure~\ref{fig:problem}. Consequently, due to the dropped tokens, the set $\mathbb{R}_i$ defined in Eq.~\eqref{eq:route-score} only includes the successfully routed experts, i.e., $|\mathbb{R}_i|$ <= k.

Conversely, certain experts may receive fewer tokens than the capacity limitation, leading to redundant computation in the form of padding. These padding instances are illustrated in yellow in Figure~\ref{fig:problem}.

If the capacity factor for top-$k$ routing is set to $k$, the number of dropped tokens and padding tokens will be equal. However, this equality does not hold if we modify the capacity factor. Increasing the capacity factor results in fewer dropped tokens but more padding. Conversely, reducing the capacity factor reduces padding tokens but increases the number of dropped tokens.

\section{Method}
In this paper, we introduce a novel approach to address both the dropped tokens and padding associated with top-$k$ routing by utilizing Rectify-Routers. Specifically, we propose two Rectify-Routers: the Intra-GPU Rectification and the Fill-in Rectification, which are visualized in Figure \ref{fig:solution}. The Intra-GPU Rectification is designed to efficiently post-process the dropped tokens, while the Fill-in Rectification is dedicated to addressing the padding problem.

\subsection{Rectify-Router for Dropped Tokens: Intra-GPU Rectification}
We expect to post-process the dropped tokens by evenly routing them across GPUs. But sending tokens among GPUs requires expensive communication cost. Furthermore, the dropped tokens have the lower routing scores than the other tokens routed to the same expert, which may be less important. Therefore, we propose an efficient Rectify-Router for the dropped tokens: Intra-GPU Rectification, which dispatch the dropped tokens to the experts inside GPU, which does not require any communication among GPUs. This process is visualized in the left part of Figure \ref{fig:solution}, where the dropped tokens from GPU 0 are routed to the expert 0 or expert 1 at GPU 0. 

Given the input token $x_i$, the Intra-GPU Rectification greedily assigns token $x_i$ to the optimal expert within the same GPU according to the routing scores. The Intra-GPU Rectification can be seen as a variant of the top-$k$ routing. If all experts are distributed in the same GPU, then the Intra-GPU Rectification is exactly the top-1 routing.

In top-$k$ routing, the same token may be dropped by multiple times. Take the top-2 routing as an example, if a token $x_i$ is dropped at both the first and second routing, it should be sent to two experts at Intra-GPU Rectification. To simplify the problem, we only send $x_i$ to one expert, although it is dropped twice. In another example, the token $x_i$ is dropped only at the second routing, while the first routing is successful. In this case, we have to combine the results of top-$k$ routing and Intra-GPU Rectification. We combine them linearly according to the routing scores:
\begin{equation}
    o_i=\frac{\sum_{j\in \mathbb{R}_i} e^{a_{ij}} {E_j(x_i)} + (k-|\mathbb{R}_i|)e^{a_{ih}}E_h(x_i)} 
    {({\sum_{j\in\mathbb{R}_i} e^{a_{ij}}})+(k-|\mathbb{R}_i|)e^{a_{ih}}},
    \label{eq:norm2}
\end{equation}
where $E_j(x_i)$ represents the expert outputs obtained through top-$k$ routing, while $E_h(x_i)$ denotes the expert outputs from Intra-GPU Rectification. We normalize the routing scores $a_{ij}$ and $a_{ih}$ as the combining weights of $E_j(x_i)$ and $E_h(x_i)$ respectively. Specifically, we scale the combining weights of $E_h(x_i)$ with a constant factor $(k-|\mathbb{R}_i|)$ , because a token is dropped $(k-|\mathbb{R}_i|)$ times but only processed by one expert in the Intra-GPU Rectification.

Similar to the top-$k$ router, the Intra-GPU Rectification also exhibits imbalance. However, this imbalance does not affect the computational fairness among GPUs. In Intra-GPU Rectification, the computation cost of a GPU is solely determined by the number of dropped tokens at that particular GPU, rather than the routing outcomes. Since the data is independently and identically distributed across devices, the number of dropped tokens on different GPUs is approximately the same. 

\subsection{Rectify-Router for Padding: Fill-in Rectification}
Fill-in Rectification aims to replace the unnecessary padding with the tokens that have high routing scores, which is visualized in the right part of Figure \ref{fig:solution}. This process is divided into two separate stages. Firstly, we identify the most suitable expert for each token, and subsequently, we select the optimal tokens for each expert.

During the initial stage, each token will choose the expert ranked as the $k+1$th highest score as the optimal expert. This decision is based on the fact that the top-$k$ experts have already been assigned, and the $k+1$th expert is considered the most suitable among the remaining experts. Furthermore, each token is only allowed to select one expert, which avoids the same token being processed by multiple experts during the second stage.

Upon completion of the first stage, we transition to the second stage. It is worth noting that multiple tokens may select the same expert. Consequently, it is possible that the number of tokens choosing a particular expert surpasses the number of padding tokens of that expert. In such scenarios, we prioritize the tokens with higher routing scores for replacing the padding tokens.

Indeed, implementing this algorithm can be achieved by extending the top-$k$ router to a top-$k+1$ router while ensuring the expert capacity remains unchanged. As the expert capacity is fixed, introducing the Fill-in Rectification incurs minimal additional overhead. Alternatively, we can view this approach as reducing the capacity factor of the top-$k+1$ routing from $k+1$ to $k$ to avoid the padding.

The Fill-in Rectification has a potential issue related to the normalization of routing scores, where the gradient of routing scores may vanish due to the invalid normalization. We address this issue in Appendix \ref{apd:grad-issue} with straight-through trick~\citep{st-trcik}.
 
\section{Experiments}
\begin{table*}[t]
    \centering
    \resizebox{\textwidth}{!}{%
    \begin{tabular}{llllllllr}
    \toprule
    \textbf{Model} & \textbf{Router}& \textbf{CF} & \textbf{Train Speed} & \textbf{MMLU} & \textbf{SuperGLUE} & \textbf{TruthfulQA} & \textbf{LogiQA}  &\textbf{Avg}\\ \midrule
    LLama2-raw& -                      & -   & 3.2k& 25.85 & 59.06 & 25.21 & 25.03  &33.78\\
    LLama2& -                      & -   & 3.2k& 35.01 & 63.74 & 30.23 & 27.64  &\textbf{39.15}\\\midrule
    \multirow{4}{*}{\shortstack{LLama-MoE\\(Top-1)}}& Top-1            & 1.0 & 2.4k& 33.05 & 64.34 & 29.49 & 28.11   &38.74\\
                           & Top-1+IR& 1.0 & 2.3k& \textbf{36.27} & 64.52 & 30.35 & \textbf{30.56}  &40.42\\
                           & Top-1+FR& 1.0 & 2.3k& 34.66 & 63.97 & 28.51 & 29.18  &39.08
    \\
                           & Top-1+FR+IR& 1.0 & 2.2k& 35.81 & \textbf{65.08} & \textbf{30.84} & \textbf{30.56}  &\textbf{40.57}\\
    \midrule
    \multirow{4}{*}{\shortstack{LLama-MoE\\(Top-2)}}& Top-2            & 2.0 & 1.7k& 35.39 & 64.58 & 29.98 & 29.33   &39.82
    \\
                           & Top-2+IR& 2.0 & 1.6k& 35.92 & 65.11 & 29.98 & 29.03  &40.01\\
                           & Top-2 + FR& 2.0 & 1.6k& 35.90 & 64.35 & \textbf{31.08} & 29.80  &40.28\\
                           & Top-2 +FR+IR& 2.0 & 1.5k& \textbf{36.01} & \textbf{65.60} & 30.72 & \textbf{29.95}  &\textbf{40.57}\\
    \bottomrule
    \end{tabular}%
    }
    \caption{The performance of LLama2-7b and MoE models on MMLU, SuperGLUE, TruthfulQA and LogiQA. CF denotes the capacity factor defined in Eq.~\eqref{capacity}. Avg represents the average accuracy. The training speed is measured as the number of tokens that each GPU can process per second. All models were trained on OpenOrca except for LLama2-raw. Top-$k$+FR and Top-$k$+IR represents the top-$k$ router using Fill-in Rectification and Intra-GPU Rectification respectively. Top-$k$+FR+IR combines both the Fill-in Rectification and the Intra-GPU Rectification. }
    \label{tab:main}
\end{table*}

\subsection{Experiment Settings}
\paragraph{Model} We follow previous work~\citep{upcycling} to train MoE models from a pretrained dense model. We initialize all experts in the same layer of MoE as the FFN parameters of the corresponding layer in the Dense model. We use the LLama2-7b~\citep{LLama2} to initialize MoE models. Following previous work~\citep{gshard,switchc},  for every alternate layer, we replaced the FFN layer in LLama2-7b with an MoE layer to transform it to an MoE model. Consequently, LLama2-7b, with 32 FFN layers, is transformed into an MoE model with 16 FFN layers and 16 MoE layers. In most of our experiments, we employ eight experts per layer in the MoE models.  But in Appendix~\ref{apd:scale-32}, we explore the extension of the number of experts to 32.  To ensure efficient computation, we evenly distribute the experts across multiple GPUs, with each GPU hosting a single expert for each layer. Our experiments are conducted using the MoE implementation of DeepSpeed~\cite{deepspeed-moe} and the training framework of gpt-neox~\citep{gpt-neox-library}.

For simplicity, we denote our Intra-GPU Rectification as \textbf{IR}, and the Fill-in Rectification as \textbf{FR}. The top-$k$ router, depending on whether it uses the Intra-GPU Rectification or the Fill-in Rectification, will be denoted as \textbf{Top-k +IR} or \textbf{Top-k+FR}, respectively.

\paragraph{Training} During the training phase, we utilize the OpenOrca dataset~\citep{OpenOrca}, which is an open-source reimplementation of Orca dataset~\citep{orca}. It augments the instructions from flan data~\citep{longpre2023flan} by adding complex system prompts and generate the step-by-step reasoning or explanation using chatgpt~\citep{chatgpt}.

We conduct our model training on a cluster of 32 GPUs (80GB). The training process consists of 10k steps with a global batch size of 256 and a micro batch size of 8. Following~\citet{orca}, we construct training examples by concatenating instructions with their corresponding responses: ``[instruct][response]''. However, only the tokens in the response are utilized for the next-token-prediction loss. To remove the padding, we fix the sequence length to be 2048 and pack multiple examples to the same sequence. For optimization, we use the Adam optimizer~\citep{adam} with an initial learning rate of 1e-5, which is decayed to 1e-6 using a cosine learning rate scheduler. Regarding the load-balance loss for the top-$k$ router, we set the weights to 1e-2, following  ~\citet{switchc}. 

\paragraph{Evaluation} We evaluated our models on multiple benchmarks, including MMLU~\citep{mmlu}, SuperGLUE~\citep{superglue}, TruthfulQA~\citep{truthfulqa} and LogiQA~\citep{logiqa}, which covers the evaluation in knowledge, natural language understanding, safety, and logical reasoning respectively. All evaluations were conducted in a zero-shot setting. Our evaluation metric was accuracy, and we utilized the lm-evaluation-harness~\citep{eval-harness} framework for conducting the evaluations.

\begin{table*}[t]
    \centering
    \resizebox{\textwidth}{!}{%
    \begin{tabular}{llllllllr}
    \toprule
    \textbf{Train Router}& \textbf{Test Router}& \textbf{Test CF} &  \textbf{Test Speed} &
    \textbf{MMLU} & \textbf{SuperGLUE} & \textbf{TruthfulQA} & \textbf{LogiQA}  &\textbf{Avg}\\ \midrule
    \multirow{4}{*}{Top-1} & Top-1                 & \multirow{4}{*}{1.0} & 9.4k & 33.05 & 64.34 & 29.49 & 28.11  &38.74\\
                           & Top-1+IR&                      & 9.2k & 36.21 & \textbf{64.64} & 29.86 & 29.18  &39.97\\
                           & Top-1+FR&                      & 8.9k & 33.28 & 62.76 & 28.51 & 29.49  &38.51
    \\
                           & Top-1+FR+IR&                   & 8.6k & \textbf{36.40}& 63.94& \textbf{29.98}& \textbf{29.80} &\textbf{40.03}\\
    \midrule
    \multirow{4}{*}{Top-2} & Top-2                 & \multirow{4}{*}{2.0} & 6.2k & 35.39 & 64.58 & 29.98 & 29.33   &39.82
    \\
                           & Top-2+IR&                      & 6.0k & 35.70 & 64.40 & \textbf{30.47} & 29.95  &40.13
    \\
                           & Top-2+FR&                      & 5.8k & 35.96 & \textbf{65.37} & 30.35 & 30.26  &40.48\\
                           & Top-(2+1)+IR&                  & 5.5k & \textbf{36.14} & 65.16& 30.35& \textbf{31.49} &\textbf{40.78}\\
    \bottomrule 
    \end{tabular}%
    }
    \caption{The performance of applying Intra-GPU Rectification and Fill-in Rectification only at inference. All models are trained with the vanilla top-1 router and top-2 router (referred to as the train router), but they were evaluated with Intra-GPU Rectification or Fill-in Rectification at inference (referred to as the test router). Test CF denotes the capacity factor set during inference. Test speed represents the number of tokens processed per second on each GPU during inference.}
    \label{tab:train-free}
\end{table*}

\subsection{Main Results}
We trained both LLama2-7b and LLama-based MoE on OpenOrca and evaluated them on MMLU (knowledge), SuperGLUE (NLU), TruthfulQA (Safety) and LogiQA (Reasoning), the results of which are shown in Table~\ref{tab:main}. Comparing the performance of LLama2-raw (pretrained) and LLama2 (trained on OpenOrca), we observed that the LLama2 outperforms LLama2-raw substantially, which demonstrates the effectiveness of finetuning on openorca. To evaluate the effectiveness of our methods, we applied our Intra-GPU Rectification (IR) and Fill-in Rectification (FR) to both the top-1 router and top-2 router. These configurations are grouped as LLama-MoE (Top-1) and LLama-MoE (Top-2) in Table~\ref{tab:main}.

\paragraph{LLama-MoE (Top-1)} We conducted 4 top-1 based MoE models (Top-1, Top-1+FR, Top-1+IR, Top-1+FR+IR). The performance of the vanilla top-1 router is subpar, and it is even inferior to the dense model (LLama2-FT) on both MMLU and TruthfulQA. But after incorporating our proposed Intra-GPU Rectification (Top-1+IR), the performance of the top-1 router are significantly improved on all benchmarks, especially on MMLU and LogiQA. This indicates that the dropped tokens have a substantial impact on model performance, and the Intra-GPU Rectification effectively handles these dropped tokens. Our Fill-in Rectification (Top-1+FR) also significantly improves the performance of the model on MMLU and LogiQA tasks. But it is worth noting that the performance of the model declined on the other two benchmarks. Therefore, it can be concluded that the primary issue with top-1 routing lies in dropped tokens rather than padding. Combing the Intra-GPU Rectification and Fill-in Rectification resulted in the best top-1-based router (Top-1+FR+IR), which outperforms the vanilla top-1 router by 1.83 (4.7\%) in terms of the average accuracy across benchmarks.

\paragraph{LLama-MoE (Top-2)} Top-2 based routers also encompass  4 routers (Top-2, Top-2-FR, Top-2-IR, Top-2-FR+IR). Both the Intra-GPU Rectification and Fill-in Rectification significantly enhance the performance of Top-2 router on at least 2 benchmarks, which demonstrate that our methods are effective for the top-2 router as well. Just as we observed with the top-1 routing results, combining the Intra-GPU Rectification and the Fill-in Rectification in the top-2 router yielded the best performance on all benchmarks. Specifically, the Top-2+FR+IR outperformed the vanilla top-2 router by a margin of 0.75 (1.8\%) in terms of the average accuracy across benchmarks.

Interestingly, we observed that the top-1 router outperformed the top-2 router in some benchmarks. For example, Top-1+FR+IR outperforms Top-1+FR+IR on both TruthfulQA and LogiQA, which raises concerns about potential overfitting in the top-2 router. Finally, it is important to note that that both the Intra-GPU Rectification and Fill-in Rectification do not alter the capacity of experts, hence they do not significantly influence the training speed.


\subsection{Improve Top-$k$ Routing at Inference}   
In this experiment, we conducted a study to evaluate the effectiveness of applying Rectify-Routers at the inference stage of MoE models. The results are presented in Table~\ref{tab:train-free}. We found that both the Intra-GPU Rectification and Fill-in Rectification can improve the performance of top-1 and top-2 routers at inference, even they are not applied at training. Similar to the results in Table~\ref{tab:main}, combining Intra-GPU Rectification and Fill-in Rectification yielded better results than using either method alone. Moreover, both the Intra-GPU Rectification and Fill-in Rectification only slightly slows down (<10\%) the inference speed of top-$k$ routers.

Comparing the results of Table~\ref{tab:main} and Table~\ref{tab:train-free}, we observed that using the Rectify-Routers (Intra-GPU Rectification and Fill-in Rectification) at both training and inference is better than only applying them at inference for top-1 based models. Conversely, for top-2 based models, the application of Rectify-Routers solely during the inference stage proves to be sufficient, as it demonstrates comparable performance to using them during both training and inference.  Therefore, we recommend training Rectify-Routers for the top-1 router and directly applying them at inference for the top-2-router.

\subsection{Capacity Factor Variation}\label{sec:capacity-var}
    
In the previous experiments, we maintained a fixed capacity factor of $k$ for top-$k$ routing, as it is a common practice. However, there are instances where it may be beneficial to adjust the capacity factor for improved efficiency or performance. Therefore, in this section, we examine the performance of our Rectify-Routers under different capacity factors. 
We anticipate that the Intra-GPU Rectification will be more effective with a lower capacity factor, as it deals with a larger number of dropped tokens. On the other hand, we expect the Fill-in Rectification to perform better with a higher capacity factor, as it introduces more padding. We validate these hypotheses in Table~\ref{tab:low-cf} and Table~\ref{tab:high-cf}, respectively. 
To minimize training costs, we train MoE models using the vanilla top-$k$ router with a capacity factor of $k$, and evaluate models with different capacity factors. We only present the average accuracy of models in Table \ref{tab:low-cf} and Table~\ref{tab:high-cf}. The complete results are shown in Appendix~\ref{app:complete}


\paragraph{Post Routing with Low Capacity} 
From Table~\ref{tab:high-cf}, we can see that decreasing the capacity factor improves the efficiency of both top-1 and top-2 based models. However, It also leads to noticeable decrease in the model performance on benchmarks. It is interesting that the top-2 router is more robust to the decrease in capacity factor. Specifically, reducing the capacity factor of the vanilla top-2 router from 2 to 1.5 only results in a slight performance decline (0.32). 

In contrast to the vanilla top-1 or top-2 routers, the MoE models incorporating our Intra-GPU Rectification (Top-1+IR and Top-2+IR) are robust to the decrease of capacity factor. We even observed that the lower capacity factor leads to a better performance for both Top-1+IR and Top-2+IR, which suggests that the Intra-GPU Rectification acts as a form of regularization for the MoE models by constraining the choices made by the experts. The similar results are also observed in~\citet{scomoe,moe-dropout}. 
By setting the capacity factor of Top-1+IR to 0.5 and that of Top-2-IR to 1.0, we observed that they are faster than the vanilla top-1 (1.13x) and top-2 routers (1.18x) respectively, while maintaining comparable or superior performance.

\begin{table}[t]
    \centering
    \resizebox{0.5\textwidth}{!}{%
    \begin{tabular}{lllll}
    \toprule
    \textbf{Train Router} & \textbf{Test Router} & \textbf{Test CF} & \textbf{Test Speed} & \textbf{Avg}\\ \midrule
    \multirow{6}{*}{Top-1} & Top-1    & \multirow{2}{*}{1.0}  & 9.4k 
    &38.74\\
                           & Top-1+IR &                       & 9.2k &\textbf{39.97}\\ \cmidrule(l){2-5} 
                           & Top-1    & \multirow{2}{*}{0.75}&  12.1k&37.83\\
                           & Top-1+IR &                       &  9.9k&\textbf{40.06}\\ \cmidrule(l){2-5} 
                           & Top-1    & \multirow{2}{*}{0.5}&  16k&34.84\\
                           & Top-1+IR &                       &  10.6k&\textbf{40.40}\\ \midrule
    \multirow{6}{*}{Top-2} & Top-2    & \multirow{2}{*}{2.0}  &  6.2k &39.82
    \\
                           & Top-2+IR &                       &  6.0k &\textbf{40.13}\\ \cmidrule(l){2-5} 
                           & Top-2    & \multirow{2}{*}{1.5}  &  7.4k&39.50\\
                           & Top-2+IR &                       &  6.6k&\textbf{39.60}\\ \cmidrule(l){2-5} 
                           & Top-2    & \multirow{2}{*}{1.0}  &  8.9k &38.51
    \\
                           & Top-2+IR &                       &  7.3k&\textbf{40.01}\\ \midrule
    \end{tabular}%
    }
    \caption{Performance of top-$k$ routers and their variants with low capacity factors ($<=k$).}
    \label{tab:low-cf}
\end{table}

\paragraph{Fill-in Rectification with High Capacity} Increasing the capacity factor of MoE models has been widely suggested in previous research studies~\citep{switchc,st-moe}. In alignment with these findings, we have also observed the benefits of increasing the capacity factor in terms of improving model performance, as demonstrated in Table~\ref{tab:high-cf}. Notably, we have found that increasing the capacity factor of the top-1 router leads to a more substantial improvement in model performance than that of the top-2 router. 

Our Fill-in Rectification introduces a more significant and consistent improvement with the increase in capacity factor. Top-1+FR and Top-2+FR consistently outperform Top-1 and Top-2, respectively, across various capacity factor settings.

Through our empirical validation, we have confirmed our hypothesis that decreasing the capacity factor benefits Intra-GPU Rectification, while increasing the capacity factor enhances the Fill-in Rectification. By combining the Intra-GPU Rectification and the Fill-in Rectification, we have achieved a model that is robust to both high and low capacity settings. 

\begin{table}[t]
    \centering
    \resizebox{0.5\textwidth}{!}{%
    \begin{tabular}{lllll}
    \toprule
    \textbf{Train Router} & \textbf{Test Router} & \textbf{Test CF}  &\textbf{Test Speed}& \textbf{Avg}\\ \midrule
    \multirow{6}{*}{Top-1} & Top-1     & \multirow{2}{*}{1.0}   &9.4k &\textbf{38.74}\\
                           & Top-1+FR&                        &8.9k&38.51
    \\\cmidrule(l){2-5} 
                           & Top-1     & \multirow{2}{*}{1.25}  &8.6k&39.59\\
                           & Top-1+FR&                        &8.1k&\textbf{40.10}\\\cmidrule(l){2-5} 
                           & Top-1     & \multirow{2}{*}{1.5}   &7.9k&39.86
    \\
                           & Top-1+FR&                        &7.3k&\textbf{40.33}\\ \midrule
    \multirow{6}{*}{Top-2} & Top-2     & \multirow{2}{*}{2.0}   &6.2k &39.82
    \\
                           & Top-2+FR&                        &5.8k &\textbf{40.48}\\ \cmidrule(l){2-5} 
                           & Top-2     & \multirow{2}{*}{2.5}   &5.4k&39.89\\
                           & Top-2+FR&                        &5.1k&\textbf{40.51}\\ \cmidrule(l){2-5} 
                           & Top-2     & \multirow{2}{*}{3.0}   &4.9k&40.03\\
                           & Top-2+FR&                        &4.5k&\textbf{40.44}\\
    \bottomrule
    \end{tabular}%
    }
    \caption{Performance of top-$k$ routers and their variants with high capacity factors ($>=k$).}
    \label{tab:high-cf}
\end{table}
\subsection{Other Experiments}
1) We scale the number of experts from 8 to 32 in Appendix ~\ref{apd:scale-32};
2) We analyze the impact of experts-GPUs distribution on our Intra-GPU Rectification in Appendix ~\ref{apd:expert-dist};
3) We validate the importance of straight-through trick in Appendix~\ref{apd:ablation-st};
4) We explore whether our methods is still effective without load-balance loss in Appendix \ref{apd:no-laux}.
\section{Conclusion}
In this paper, we present the Rectify-Router, a method to tackle dropped tokens and padding in MoE models. By introducing the Intra-GPU Rectification and the Fill-in Rectification, we effectively handle the issues of dropped tokens and padding, respectively. Experimental results demonstrate the individual effectiveness of both techniques and the synergistic performance improvement when they are combined. Furthermore, our methods prove to be effective in diverse settings, including varying numbers of experts, different expert capacities, and even without the load-balance loss. 

\section{Limitation}
In this paper, we propose Rectify-Router to tackle the issues of dropped tokens and padding in the top-$k$ router. The effectiveness of our methods has been demonstrated through experiments. But our experiments still have the following limitations due to the expensive training cost:
\begin{enumerate}
    \item The MoE models trained in this work are initialized from a dense model (LLama2-7b). We have not validated our methods by training from scratch.
    \item Our experiments are conducted based on LLama2-7b, while the other settings like LLama2-70B have not been explored.
\end{enumerate}

These limitations highlight potential areas for future research and expansion of our work.
\bibliography{acl_latex}

\begin{thebibliography}{34}
\expandafter\ifx\csname natexlab\endcsname\relax\def\natexlab#1{#1}\fi

\bibitem[{Andonian et~al.(2021)Andonian, Anthony, Biderman, Black, Gali, Gao, Hallahan, Levy-Kramer, Leahy, Nestler, Parker, Pieler, Purohit, Songz, Phil, and Weinbach}]{gpt-neox-library}
Alex Andonian, Quentin Anthony, Stella Biderman, Sid Black, Preetham Gali, Leo Gao, Eric Hallahan, Josh Levy-Kramer, Connor Leahy, Lucas Nestler, Kip Parker, Michael Pieler, Shivanshu Purohit, Tri Songz, Wang Phil, and Samuel Weinbach. 2021.
\newblock \href {https://doi.org/10.5281/zenodo.5879544} {{GPT-NeoX: Large Scale Autoregressive Language Modeling in PyTorch}}.

\bibitem[{Bengio et~al.(2013)Bengio, L{\'{e}}onard, and Courville}]{st-trcik}
Yoshua Bengio, Nicholas L{\'{e}}onard, and Aaron~C. Courville. 2013.
\newblock \href {http://arxiv.org/abs/1308.3432} {Estimating or propagating gradients through stochastic neurons for conditional computation}.
\newblock \emph{CoRR}, abs/1308.3432.

\bibitem[{Dai et~al.(2024)Dai, Deng, Zhao, Xu, Gao, Chen, Li, Zeng, Yu, Wu, Xie, Li, Huang, Luo, Ruan, Sui, and Liang}]{deepseek-moe}
Damai Dai, Chengqi Deng, Chenggang Zhao, R.~X. Xu, Huazuo Gao, Deli Chen, Jiashi Li, Wangding Zeng, Xingkai Yu, Y.~Wu, Zhenda Xie, Y.~K. Li, Panpan Huang, Fuli Luo, Chong Ruan, Zhifang Sui, and Wenfeng Liang. 2024.
\newblock \href {https://doi.org/10.48550/ARXIV.2401.06066} {Deepseekmoe: Towards ultimate expert specialization in mixture-of-experts language models}.
\newblock \emph{CoRR}, abs/2401.06066.

\bibitem[{Du et~al.(2022)Du, Huang, Dai, Tong, Lepikhin, Xu, Krikun, Zhou, Yu, Firat, Zoph, Fedus, Bosma, Zhou, Wang, Wang, Webster, Pellat, Robinson, Meier{-}Hellstern, Duke, Dixon, Zhang, Le, Wu, Chen, and Cui}]{glam}
Nan Du, Yanping Huang, Andrew~M. Dai, Simon Tong, Dmitry Lepikhin, Yuanzhong Xu, Maxim Krikun, Yanqi Zhou, Adams~Wei Yu, Orhan Firat, Barret Zoph, Liam Fedus, Maarten~P. Bosma, Zongwei Zhou, Tao Wang, Yu~Emma Wang, Kellie Webster, Marie Pellat, Kevin Robinson, Kathleen~S. Meier{-}Hellstern, Toju Duke, Lucas Dixon, Kun Zhang, Quoc~V. Le, Yonghui Wu, Zhifeng Chen, and Claire Cui. 2022.
\newblock \href {https://proceedings.mlr.press/v162/du22c.html} {Glam: Efficient scaling of language models with mixture-of-experts}.
\newblock In \emph{International Conference on Machine Learning, {ICML} 2022, 17-23 July 2022, Baltimore, Maryland, {USA}}, volume 162 of \emph{Proceedings of Machine Learning Research}, pages 5547--5569. {PMLR}.

\bibitem[{Fedus et~al.(2022)Fedus, Zoph, and Shazeer}]{switchc}
William Fedus, Barret Zoph, and Noam Shazeer. 2022.
\newblock \href {http://jmlr.org/papers/v23/21-0998.html} {Switch transformers: Scaling to trillion parameter models with simple and efficient sparsity}.
\newblock \emph{J. Mach. Learn. Res.}, 23:120:1--120:39.

\bibitem[{Gale et~al.(2022)Gale, Narayanan, Young, and Zaharia}]{megablock}
Trevor Gale, Deepak Narayanan, Cliff Young, and Matei Zaharia. 2022.
\newblock \href {https://doi.org/10.48550/ARXIV.2211.15841} {Megablocks: Efficient sparse training with mixture-of-experts}.
\newblock \emph{CoRR}, abs/2211.15841.

\bibitem[{Gao et~al.(2023)Gao, Tow, Abbasi, Biderman, Black, DiPofi, Foster, Golding, Hsu, Le~Noac'h, Li, McDonell, Muennighoff, Ociepa, Phang, Reynolds, Schoelkopf, Skowron, Sutawika, Tang, Thite, Wang, Wang, and Zou}]{eval-harness}
Leo Gao, Jonathan Tow, Baber Abbasi, Stella Biderman, Sid Black, Anthony DiPofi, Charles Foster, Laurence Golding, Jeffrey Hsu, Alain Le~Noac'h, Haonan Li, Kyle McDonell, Niklas Muennighoff, Chris Ociepa, Jason Phang, Laria Reynolds, Hailey Schoelkopf, Aviya Skowron, Lintang Sutawika, Eric Tang, Anish Thite, Ben Wang, Kevin Wang, and Andy Zou. 2023.
\newblock \href {https://doi.org/10.5281/zenodo.10256836} {A framework for few-shot language model evaluation}.

\bibitem[{Jiang et~al.(2024)Jiang, Sablayrolles, Roux, Mensch, Savary, Bamford, Chaplot, de~Las~Casas, Hanna, Bressand, Lengyel, Bour, Lample, Lavaud, Saulnier, Lachaux, Stock, Subramanian, Yang, Antoniak, Scao, Gervet, Lavril, Wang, Lacroix, and Sayed}]{mixtral}
Albert~Q. Jiang, Alexandre Sablayrolles, Antoine Roux, Arthur Mensch, Blanche Savary, Chris Bamford, Devendra~Singh Chaplot, Diego de~Las~Casas, Emma~Bou Hanna, Florian Bressand, Gianna Lengyel, Guillaume Bour, Guillaume Lample, L{\'{e}}lio~Renard Lavaud, Lucile Saulnier, Marie{-}Anne Lachaux, Pierre Stock, Sandeep Subramanian, Sophia Yang, Szymon Antoniak, Teven~Le Scao, Th{\'{e}}ophile Gervet, Thibaut Lavril, Thomas Wang, Timoth{\'{e}}e Lacroix, and William~El Sayed. 2024.
\newblock \href {https://doi.org/10.48550/ARXIV.2401.04088} {Mixtral of experts}.
\newblock \emph{CoRR}, abs/2401.04088.

\bibitem[{Kingma and Ba(2015)}]{adam}
Diederik~P. Kingma and Jimmy Ba. 2015.
\newblock \href {http://arxiv.org/abs/1412.6980} {Adam: {A} method for stochastic optimization}.
\newblock In \emph{3rd International Conference on Learning Representations, {ICLR} 2015, San Diego, CA, USA, May 7-9, 2015, Conference Track Proceedings}.

\bibitem[{Koishekenov et~al.(2022)Koishekenov, Nikoulina, and Berard}]{nllb}
Yeskendir Koishekenov, Vassilina Nikoulina, and Alexandre Berard. 2022.
\newblock \href {https://doi.org/10.48550/ARXIV.2212.09811} {Memory-efficient {NLLB-200:} language-specific expert pruning of a massively multilingual machine translation model}.
\newblock \emph{CoRR}, abs/2212.09811.

\bibitem[{Komatsuzaki et~al.(2023)Komatsuzaki, Puigcerver, Lee{-}Thorp, Ruiz, Mustafa, Ainslie, Tay, Dehghani, and Houlsby}]{upcycling}
Aran Komatsuzaki, Joan Puigcerver, James Lee{-}Thorp, Carlos~Riquelme Ruiz, Basil Mustafa, Joshua Ainslie, Yi~Tay, Mostafa Dehghani, and Neil Houlsby. 2023.
\newblock \href {https://openreview.net/pdf?id=T5nUQDrM4u} {Sparse upcycling: Training mixture-of-experts from dense checkpoints}.
\newblock In \emph{The Eleventh International Conference on Learning Representations, {ICLR} 2023, Kigali, Rwanda, May 1-5, 2023}. OpenReview.net.

\bibitem[{Lepikhin et~al.(2021)Lepikhin, Lee, Xu, Chen, Firat, Huang, Krikun, Shazeer, and Chen}]{gshard}
Dmitry Lepikhin, HyoukJoong Lee, Yuanzhong Xu, Dehao Chen, Orhan Firat, Yanping Huang, Maxim Krikun, Noam Shazeer, and Zhifeng Chen. 2021.
\newblock \href {https://openreview.net/forum?id=qrwe7XHTmYb} {Gshard: Scaling giant models with conditional computation and automatic sharding}.
\newblock In \emph{9th International Conference on Learning Representations, {ICLR} 2021, Virtual Event, Austria, May 3-7, 2021}. OpenReview.net.

\bibitem[{Lewis et~al.(2021)Lewis, Bhosale, Dettmers, Goyal, and Zettlemoyer}]{base-layer}
Mike Lewis, Shruti Bhosale, Tim Dettmers, Naman Goyal, and Luke Zettlemoyer. 2021.
\newblock \href {http://proceedings.mlr.press/v139/lewis21a.html} {{BASE} layers: Simplifying training of large, sparse models}.
\newblock In \emph{Proceedings of the 38th International Conference on Machine Learning, {ICML} 2021, 18-24 July 2021, Virtual Event}, volume 139 of \emph{Proceedings of Machine Learning Research}, pages 6265--6274. {PMLR}.

\bibitem[{Li et~al.(2023)Li, Zhang, Koto, Yang, Zhao, Gong, Duan, and Baldwin}]{mmlu}
Haonan Li, Yixuan Zhang, Fajri Koto, Yifei Yang, Hai Zhao, Yeyun Gong, Nan Duan, and Timothy Baldwin. 2023.
\newblock \href {https://doi.org/10.48550/ARXIV.2306.09212} {{CMMLU:} measuring massive multitask language understanding in chinese}.
\newblock \emph{CoRR}, abs/2306.09212.

\bibitem[{Lian et~al.(2023)Lian, Goodson, Pentland, Cook, Vong, and "Teknium"}]{OpenOrca}
Wing Lian, Bleys Goodson, Eugene Pentland, Austin Cook, Chanvichet Vong, and "Teknium". 2023.
\newblock Openorca: An open dataset of gpt augmented flan reasoning traces.
\newblock \url{https://https://huggingface.co/Open-Orca/OpenOrca}.

\bibitem[{Lin et~al.(2022)Lin, Hilton, and Evans}]{truthfulqa}
Stephanie Lin, Jacob Hilton, and Owain Evans. 2022.
\newblock \href {https://doi.org/10.18653/V1/2022.ACL-LONG.229} {Truthfulqa: Measuring how models mimic human falsehoods}.
\newblock In \emph{Proceedings of the 60th Annual Meeting of the Association for Computational Linguistics (Volume 1: Long Papers), {ACL} 2022, Dublin, Ireland, May 22-27, 2022}, pages 3214--3252. Association for Computational Linguistics.

\bibitem[{Liu et~al.(2020)Liu, Cui, Liu, Huang, Wang, and Zhang}]{logiqa}
Jian Liu, Leyang Cui, Hanmeng Liu, Dandan Huang, Yile Wang, and Yue Zhang. 2020.
\newblock \href {https://doi.org/10.24963/IJCAI.2020/501} {Logiqa: {A} challenge dataset for machine reading comprehension with logical reasoning}.
\newblock In \emph{Proceedings of the Twenty-Ninth International Joint Conference on Artificial Intelligence, {IJCAI} 2020}, pages 3622--3628. ijcai.org.

\bibitem[{Liu et~al.(2022)Liu, Kim, Muzio, and Hassan}]{moe-dropout}
Rui Liu, Young~Jin Kim, Alexandre Muzio, and Hany Hassan. 2022.
\newblock \href {https://proceedings.mlr.press/v162/liu22g.html} {Gating dropout: Communication-efficient regularization for sparsely activated transformers}.
\newblock In \emph{International Conference on Machine Learning, {ICML} 2022, 17-23 July 2022, Baltimore, Maryland, {USA}}, volume 162 of \emph{Proceedings of Machine Learning Research}, pages 13782--13792. {PMLR}.

\bibitem[{Longpre et~al.(2023)Longpre, Hou, Vu, Webson, Chung, Tay, Zhou, Le, Zoph, Wei, and Roberts}]{longpre2023flan}
Shayne Longpre, Le~Hou, Tu~Vu, Albert Webson, Hyung~Won Chung, Yi~Tay, Denny Zhou, Quoc~V. Le, Barret Zoph, Jason Wei, and Adam Roberts. 2023.
\newblock \href {http://arxiv.org/abs/2301.13688} {The flan collection: Designing data and methods for effective instruction tuning}.

\bibitem[{Mukherjee et~al.(2023)Mukherjee, Mitra, Jawahar, Agarwal, Palangi, and Awadallah}]{orca}
Subhabrata Mukherjee, Arindam Mitra, Ganesh Jawahar, Sahaj Agarwal, Hamid Palangi, and Ahmed Awadallah. 2023.
\newblock \href {https://doi.org/10.48550/ARXIV.2306.02707} {Orca: Progressive learning from complex explanation traces of {GPT-4}}.
\newblock \emph{CoRR}, abs/2306.02707.

\bibitem[{Mustafa et~al.(2022)Mustafa, Riquelme, Puigcerver, Jenatton, and Houlsby}]{mmoe}
Basil Mustafa, Carlos Riquelme, Joan Puigcerver, Rodolphe Jenatton, and Neil Houlsby. 2022.
\newblock \href {http://papers.nips.cc/paper\_files/paper/2022/hash/3e67e84abf900bb2c7cbd5759bfce62d-Abstract-Conference.html} {Multimodal contrastive learning with limoe: the language-image mixture of experts}.
\newblock In \emph{Advances in Neural Information Processing Systems 35: Annual Conference on Neural Information Processing Systems 2022, NeurIPS 2022, New Orleans, LA, USA, November 28 - December 9, 2022}.

\bibitem[{OpenAI et~al.(2023)OpenAI, :, Achiam, Adler, Agarwal, Ahmad, Akkaya, Aleman, Almeida, Altenschmidt, Altman, Anadkat, Avila, Babuschkin, Balaji, Balcom, Baltescu, Bao, Bavarian, Belgum, Bello, Berdine, Bernadett-Shapiro, Berner, Bogdonoff, Boiko, Boyd, Brakman, Brockman, Brooks, Brundage, Button, Cai, Campbell, Cann, Carey, Carlson, Carmichael, Chan, Chang, Chantzis, Chen, Chen, Chen, Chen, Chen, Chess, Cho, Chu, Chung, Cummings, Currier, Dai, Decareaux, Degry, Deutsch, Deville, Dhar, Dohan, Dowling, Dunning, Ecoffet, Eleti, Eloundou, Farhi, Fedus, Felix, Fishman, Forte, Fulford, Gao, Georges, Gibson, Goel, Gogineni, Goh, Gontijo-Lopes, Gordon, Grafstein, Gray, Greene, Gross, Gu, Guo, Hallacy, Han, Harris, He, Heaton, Heidecke, Hesse, Hickey, Hickey, Hoeschele, Houghton, Hsu, Hu, Hu, Huizinga, Jain, Jain, Jang, Jiang, Jiang, Jin, Jin, Jomoto, Jonn, Jun, Kaftan, Łukasz Kaiser, Kamali, Kanitscheider, Keskar, Khan, Kilpatrick, Kim, Kim, Kim, Kirchner, Kiros, Knight, Kokotajlo, Łukasz Kondraciuk,
  Kondrich, Konstantinidis, Kosic, Krueger, Kuo, Lampe, Lan, Lee, Leike, Leung, Levy, Li, Lim, Lin, Lin, Litwin, Lopez, Lowe, Lue, Makanju, Malfacini, Manning, Markov, Markovski, Martin, Mayer, Mayne, McGrew, McKinney, McLeavey, McMillan, McNeil, Medina, Mehta, Menick, Metz, Mishchenko, Mishkin, Monaco, Morikawa, Mossing, Mu, Murati, Murk, Mély, Nair, Nakano, Nayak, Neelakantan, Ngo, Noh, Ouyang, O'Keefe, Pachocki, Paino, Palermo, Pantuliano, Parascandolo, Parish, Parparita, Passos, Pavlov, Peng, Perelman, de~Avila Belbute~Peres, Petrov, de~Oliveira~Pinto, Michael, Pokorny, Pokrass, Pong, Powell, Power, Power, Proehl, Puri, Radford, Rae, Ramesh, Raymond, Real, Rimbach, Ross, Rotsted, Roussez, Ryder, Saltarelli, Sanders, Santurkar, Sastry, Schmidt, Schnurr, Schulman, Selsam, Sheppard, Sherbakov, Shieh, Shoker, Shyam, Sidor, Sigler, Simens, Sitkin, Slama, Sohl, Sokolowsky, Song, Staudacher, Such, Summers, Sutskever, Tang, Tezak, Thompson, Tillet, Tootoonchian, Tseng, Tuggle, Turley, Tworek, Uribe, Vallone,
  Vijayvergiya, Voss, Wainwright, Wang, Wang, Wang, Ward, Wei, Weinmann, Welihinda, Welinder, Weng, Weng, Wiethoff, Willner, Winter, Wolrich, Wong, Workman, Wu, Wu, Wu, Xiao, Xu, Yoo, Yu, Yuan, Zaremba, Zellers, Zhang, Zhang, Zhao, Zheng, Zhuang, Zhuk, and Zoph}]{chatgpt}
OpenAI, :, Josh Achiam, Steven Adler, Sandhini Agarwal, Lama Ahmad, Ilge Akkaya, Florencia~Leoni Aleman, Diogo Almeida, Janko Altenschmidt, Sam Altman, Shyamal Anadkat, Red Avila, Igor Babuschkin, Suchir Balaji, Valerie Balcom, Paul Baltescu, Haiming Bao, Mo~Bavarian, Jeff Belgum, Irwan Bello, Jake Berdine, Gabriel Bernadett-Shapiro, Christopher Berner, Lenny Bogdonoff, Oleg Boiko, Madelaine Boyd, Anna-Luisa Brakman, Greg Brockman, Tim Brooks, Miles Brundage, Kevin Button, Trevor Cai, Rosie Campbell, Andrew Cann, Brittany Carey, Chelsea Carlson, Rory Carmichael, Brooke Chan, Che Chang, Fotis Chantzis, Derek Chen, Sully Chen, Ruby Chen, Jason Chen, Mark Chen, Ben Chess, Chester Cho, Casey Chu, Hyung~Won Chung, Dave Cummings, Jeremiah Currier, Yunxing Dai, Cory Decareaux, Thomas Degry, Noah Deutsch, Damien Deville, Arka Dhar, David Dohan, Steve Dowling, Sheila Dunning, Adrien Ecoffet, Atty Eleti, Tyna Eloundou, David Farhi, Liam Fedus, Niko Felix, Simón~Posada Fishman, Juston Forte, Isabella Fulford, Leo Gao,
  Elie Georges, Christian Gibson, Vik Goel, Tarun Gogineni, Gabriel Goh, Rapha Gontijo-Lopes, Jonathan Gordon, Morgan Grafstein, Scott Gray, Ryan Greene, Joshua Gross, Shixiang~Shane Gu, Yufei Guo, Chris Hallacy, Jesse Han, Jeff Harris, Yuchen He, Mike Heaton, Johannes Heidecke, Chris Hesse, Alan Hickey, Wade Hickey, Peter Hoeschele, Brandon Houghton, Kenny Hsu, Shengli Hu, Xin Hu, Joost Huizinga, Shantanu Jain, Shawn Jain, Joanne Jang, Angela Jiang, Roger Jiang, Haozhun Jin, Denny Jin, Shino Jomoto, Billie Jonn, Heewoo Jun, Tomer Kaftan, Łukasz Kaiser, Ali Kamali, Ingmar Kanitscheider, Nitish~Shirish Keskar, Tabarak Khan, Logan Kilpatrick, Jong~Wook Kim, Christina Kim, Yongjik Kim, Hendrik Kirchner, Jamie Kiros, Matt Knight, Daniel Kokotajlo, Łukasz Kondraciuk, Andrew Kondrich, Aris Konstantinidis, Kyle Kosic, Gretchen Krueger, Vishal Kuo, Michael Lampe, Ikai Lan, Teddy Lee, Jan Leike, Jade Leung, Daniel Levy, Chak~Ming Li, Rachel Lim, Molly Lin, Stephanie Lin, Mateusz Litwin, Theresa Lopez, Ryan Lowe,
  Patricia Lue, Anna Makanju, Kim Malfacini, Sam Manning, Todor Markov, Yaniv Markovski, Bianca Martin, Katie Mayer, Andrew Mayne, Bob McGrew, Scott~Mayer McKinney, Christine McLeavey, Paul McMillan, Jake McNeil, David Medina, Aalok Mehta, Jacob Menick, Luke Metz, Andrey Mishchenko, Pamela Mishkin, Vinnie Monaco, Evan Morikawa, Daniel Mossing, Tong Mu, Mira Murati, Oleg Murk, David Mély, Ashvin Nair, Reiichiro Nakano, Rajeev Nayak, Arvind Neelakantan, Richard Ngo, Hyeonwoo Noh, Long Ouyang, Cullen O'Keefe, Jakub Pachocki, Alex Paino, Joe Palermo, Ashley Pantuliano, Giambattista Parascandolo, Joel Parish, Emy Parparita, Alex Passos, Mikhail Pavlov, Andrew Peng, Adam Perelman, Filipe de~Avila Belbute~Peres, Michael Petrov, Henrique~Ponde de~Oliveira~Pinto, Michael, Pokorny, Michelle Pokrass, Vitchyr Pong, Tolly Powell, Alethea Power, Boris Power, Elizabeth Proehl, Raul Puri, Alec Radford, Jack Rae, Aditya Ramesh, Cameron Raymond, Francis Real, Kendra Rimbach, Carl Ross, Bob Rotsted, Henri Roussez, Nick Ryder,
  Mario Saltarelli, Ted Sanders, Shibani Santurkar, Girish Sastry, Heather Schmidt, David Schnurr, John Schulman, Daniel Selsam, Kyla Sheppard, Toki Sherbakov, Jessica Shieh, Sarah Shoker, Pranav Shyam, Szymon Sidor, Eric Sigler, Maddie Simens, Jordan Sitkin, Katarina Slama, Ian Sohl, Benjamin Sokolowsky, Yang Song, Natalie Staudacher, Felipe~Petroski Such, Natalie Summers, Ilya Sutskever, Jie Tang, Nikolas Tezak, Madeleine Thompson, Phil Tillet, Amin Tootoonchian, Elizabeth Tseng, Preston Tuggle, Nick Turley, Jerry Tworek, Juan Felipe~Cerón Uribe, Andrea Vallone, Arun Vijayvergiya, Chelsea Voss, Carroll Wainwright, Justin~Jay Wang, Alvin Wang, Ben Wang, Jonathan Ward, Jason Wei, CJ~Weinmann, Akila Welihinda, Peter Welinder, Jiayi Weng, Lilian Weng, Matt Wiethoff, Dave Willner, Clemens Winter, Samuel Wolrich, Hannah Wong, Lauren Workman, Sherwin Wu, Jeff Wu, Michael Wu, Kai Xiao, Tao Xu, Sarah Yoo, Kevin Yu, Qiming Yuan, Wojciech Zaremba, Rowan Zellers, Chong Zhang, Marvin Zhang, Shengjia Zhao, Tianhao
  Zheng, Juntang Zhuang, William Zhuk, and Barret Zoph. 2023.
\newblock \href {http://arxiv.org/abs/2303.08774} {Gpt-4 technical report}.

\bibitem[{Ott et~al.(2019)Ott, Edunov, Baevski, Fan, Gross, Ng, Grangier, and Auli}]{fairseq}
Myle Ott, Sergey Edunov, Alexei Baevski, Angela Fan, Sam Gross, Nathan Ng, David Grangier, and Michael Auli. 2019.
\newblock \href {https://doi.org/10.18653/V1/N19-4009} {fairseq: {A} fast, extensible toolkit for sequence modeling}.
\newblock In \emph{Proceedings of the 2019 Conference of the North American Chapter of the Association for Computational Linguistics: Human Language Technologies, {NAACL-HLT} 2019, Minneapolis, MN, USA, June 2-7, 2019, Demonstrations}, pages 48--53. Association for Computational Linguistics.

\bibitem[{Puigcerver et~al.(2023)Puigcerver, Riquelme, Mustafa, and Houlsby}]{soft-moe}
Joan Puigcerver, Carlos Riquelme, Basil Mustafa, and Neil Houlsby. 2023.
\newblock \href {https://doi.org/10.48550/ARXIV.2308.00951} {From sparse to soft mixtures of experts}.
\newblock \emph{CoRR}, abs/2308.00951.

\bibitem[{Rajbhandari et~al.(2022)Rajbhandari, Li, Yao, Zhang, Aminabadi, Awan, Rasley, and He}]{deepspeed-moe}
Samyam Rajbhandari, Conglong Li, Zhewei Yao, Minjia Zhang, Reza~Yazdani Aminabadi, Ammar~Ahmad Awan, Jeff Rasley, and Yuxiong He. 2022.
\newblock \href {https://proceedings.mlr.press/v162/rajbhandari22a.html} {Deepspeed-moe: Advancing mixture-of-experts inference and training to power next-generation {AI} scale}.
\newblock In \emph{International Conference on Machine Learning, {ICML} 2022, 17-23 July 2022, Baltimore, Maryland, {USA}}, volume 162 of \emph{Proceedings of Machine Learning Research}, pages 18332--18346. {PMLR}.

\bibitem[{Shazeer et~al.(2017)Shazeer, Mirhoseini, Maziarz, Davis, Le, Hinton, and Dean}]{first-moe}
Noam Shazeer, Azalia Mirhoseini, Krzysztof Maziarz, Andy Davis, Quoc~V. Le, Geoffrey~E. Hinton, and Jeff Dean. 2017.
\newblock \href {https://openreview.net/forum?id=B1ckMDqlg} {Outrageously large neural networks: The sparsely-gated mixture-of-experts layer}.
\newblock In \emph{5th International Conference on Learning Representations, {ICLR} 2017, Toulon, France, April 24-26, 2017, Conference Track Proceedings}. OpenReview.net.

\bibitem[{Shen et~al.(2023)Shen, Hou, Zhou, Du, Longpre, Wei, Chung, Zoph, Fedus, Chen, Vu, Wu, Chen, Webson, Li, Zhao, Yu, Keutzer, Darrell, and Zhou}]{flan-moe}
Sheng Shen, Le~Hou, Yanqi Zhou, Nan Du, Shayne Longpre, Jason Wei, Hyung~Won Chung, Barret Zoph, William Fedus, Xinyun Chen, Tu~Vu, Yuexin Wu, Wuyang Chen, Albert Webson, Yunxuan Li, Vincent Zhao, Hongkun Yu, Kurt Keutzer, Trevor Darrell, and Denny Zhou. 2023.
\newblock \href {http://arxiv.org/abs/2305.14705} {Mixture-of-experts meets instruction tuning:a winning combination for large language models}.

\bibitem[{Touvron et~al.(2023)Touvron, Martin, Stone, Albert, Almahairi, Babaei, Bashlykov, Batra, Bhargava, Bhosale, Bikel, Blecher, Canton{-}Ferrer, Chen, Cucurull, Esiobu, Fernandes, Fu, Fu, Fuller, Gao, Goswami, Goyal, Hartshorn, Hosseini, Hou, Inan, Kardas, Kerkez, Khabsa, Kloumann, Korenev, Koura, Lachaux, Lavril, Lee, Liskovich, Lu, Mao, Martinet, Mihaylov, Mishra, Molybog, Nie, Poulton, Reizenstein, Rungta, Saladi, Schelten, Silva, Smith, Subramanian, Tan, Tang, Taylor, Williams, Kuan, Xu, Yan, Zarov, Zhang, Fan, Kambadur, Narang, Rodriguez, Stojnic, Edunov, and Scialom}]{LLama2}
Hugo Touvron, Louis Martin, Kevin Stone, Peter Albert, Amjad Almahairi, Yasmine Babaei, Nikolay Bashlykov, Soumya Batra, Prajjwal Bhargava, Shruti Bhosale, Dan Bikel, Lukas Blecher, Cristian Canton{-}Ferrer, Moya Chen, Guillem Cucurull, David Esiobu, Jude Fernandes, Jeremy Fu, Wenyin Fu, Brian Fuller, Cynthia Gao, Vedanuj Goswami, Naman Goyal, Anthony Hartshorn, Saghar Hosseini, Rui Hou, Hakan Inan, Marcin Kardas, Viktor Kerkez, Madian Khabsa, Isabel Kloumann, Artem Korenev, Punit~Singh Koura, Marie{-}Anne Lachaux, Thibaut Lavril, Jenya Lee, Diana Liskovich, Yinghai Lu, Yuning Mao, Xavier Martinet, Todor Mihaylov, Pushkar Mishra, Igor Molybog, Yixin Nie, Andrew Poulton, Jeremy Reizenstein, Rashi Rungta, Kalyan Saladi, Alan Schelten, Ruan Silva, Eric~Michael Smith, Ranjan Subramanian, Xiaoqing~Ellen Tan, Binh Tang, Ross Taylor, Adina Williams, Jian~Xiang Kuan, Puxin Xu, Zheng Yan, Iliyan Zarov, Yuchen Zhang, Angela Fan, Melanie Kambadur, Sharan Narang, Aur{\'{e}}lien Rodriguez, Robert Stojnic, Sergey Edunov,
  and Thomas Scialom. 2023.
\newblock \href {https://doi.org/10.48550/ARXIV.2307.09288} {Llama 2: Open foundation and fine-tuned chat models}.
\newblock \emph{CoRR}, abs/2307.09288.

\bibitem[{Wang et~al.(2019)Wang, Pruksachatkun, Nangia, Singh, Michael, Hill, Levy, and Bowman}]{superglue}
Alex Wang, Yada Pruksachatkun, Nikita Nangia, Amanpreet Singh, Julian Michael, Felix Hill, Omer Levy, and Samuel~R. Bowman. 2019.
\newblock \href {https://proceedings.neurips.cc/paper/2019/hash/4496bf24afe7fab6f046bf4923da8de6-Abstract.html} {Superglue: {A} stickier benchmark for general-purpose language understanding systems}.
\newblock In \emph{Advances in Neural Information Processing Systems 32: Annual Conference on Neural Information Processing Systems 2019, NeurIPS 2019, December 8-14, 2019, Vancouver, BC, Canada}, pages 3261--3275.

\bibitem[{Yu et~al.(2022)Yu, Artetxe, Ott, Shleifer, Gong, Stoyanov, and Li}]{mlp-moe}
Ping Yu, Mikel Artetxe, Myle Ott, Sam Shleifer, Hongyu Gong, Ves Stoyanov, and Xian Li. 2022.
\newblock \href {https://doi.org/10.48550/ARXIV.2203.06850} {Efficient language modeling with sparse all-mlp}.
\newblock \emph{CoRR}, abs/2203.06850.

\bibitem[{Zeng and Xiong(2023)}]{scomoe}
Zhiyuan Zeng and Deyi Xiong. 2023.
\newblock \href {https://openreview.net/pdf?id=s-c96mSU0u5} {Scomoe: Efficient mixtures of experts with structured communication}.
\newblock In \emph{The Eleventh International Conference on Learning Representations, {ICLR} 2023, Kigali, Rwanda, May 1-5, 2023}. OpenReview.net.

\bibitem[{Zhou et~al.(2022)Zhou, Lei, Liu, Du, Huang, Zhao, Dai, Chen, Le, and Laudon}]{expert-choices}
Yanqi Zhou, Tao Lei, Hanxiao Liu, Nan Du, Yanping Huang, Vincent Zhao, Andrew~M. Dai, Zhifeng Chen, Quoc~V. Le, and James Laudon. 2022.
\newblock \href {http://papers.nips.cc/paper\_files/paper/2022/hash/2f00ecd787b432c1d36f3de9800728eb-Abstract-Conference.html} {Mixture-of-experts with expert choice routing}.
\newblock In \emph{Advances in Neural Information Processing Systems 35: Annual Conference on Neural Information Processing Systems 2022, NeurIPS 2022, New Orleans, LA, USA, November 28 - December 9, 2022}.

\bibitem[{Zoph et~al.(2022)Zoph, Bello, Kumar, Du, Huang, Dean, Shazeer, and Fedus}]{st-moe}
Barret Zoph, Irwan Bello, Sameer Kumar, Nan Du, Yanping Huang, Jeff Dean, Noam Shazeer, and William Fedus. 2022.
\newblock \href {http://arxiv.org/abs/2202.08906} {St-moe: Designing stable and transferable sparse expert models}.

\bibitem[{Zuo et~al.(2022)Zuo, Liu, Jiao, Kim, Hassan, Zhang, Gao, and Zhao}]{random-moe}
Simiao Zuo, Xiaodong Liu, Jian Jiao, Young~Jin Kim, Hany Hassan, Ruofei Zhang, Jianfeng Gao, and Tuo Zhao. 2022.
\newblock \href {https://openreview.net/forum?id=B72HXs80q4} {Taming sparsely activated transformer with stochastic experts}.
\newblock In \emph{The Tenth International Conference on Learning Representations, {ICLR} 2022, Virtual Event, April 25-29, 2022}. OpenReview.net.

\end{thebibliography}
\bibliographystyle{acl_natbib}
\newpage
\appendix
\section{Gradient Issues in Fill-in Rectification} \label{apd:grad-issue}
There is a potential issue in the Fill-in Rectification, which stems from the implementation of the top-$k$ routing. According to Eq.~\eqref{eq:route-score}, the routing scores of top-$k$ routing are normalized on the selected experts $\mathbb{R}_i$, rather than considering all expert choices. Several implementations like deepspeed-moe~\cite{deepspeed-moe} and fairseq-moe~\cite{fairseq} first normalize the routing scores on all experts and then re-normalize the scores specifically for the selected experts:
 
\begin{align}
g_{ij}&=\frac{e^{a_{ij}}}{\sum_j^m e^{a_{ij}}}\label{eq:norm-deepspeed}\\
    o_i&=\sum_{j\in \mathbb{R}_i} \frac{g_{ij}}{\sum_j g_{ij}} {E_j(x_i)}, \notag
\end{align}

where $g_{ij}$ represents the routing scores that are initially normalized across all experts and then further normalized specifically on the selected experts ($\mathbb{R}_i$).  However, their implementation is equivalent to directly normalizing the routing scores on $\mathbb{R}_i$. 

There are two potential issue of normalizing routing scores on $\mathbb{R}_i$: 
1) the routing scores of activated experts can not influence those of inactivated experts. For example, the increase of $a_{ij}(j\in\mathbb{R}_i)$ does not lead to the decrease of $a_{il}(l \notin \mathbb{R}_i)$ . 
2) In the case of top-2 routing, if the first routing of $x_i$ is successful while the second routing fails due to the expert overflow, the gradients of all routing scores of $x_i$ will be zero ($\frac{\partial L}{\partial a_{ij}}=0$). This is because that there is only one available expert choice for $x_i$ ($ |\mathbb{R}_i|=1$). Normalizing on $ |\mathbb{R}_i|$ would always yield a value of 1, regardless of the actual value of $a_{ij}$, leading to invalid gradients.

This problem is more prominent for the Fill-in Rectification, since it brings more dropped tokens, i.e., more unsuccessful routing. To address this problem, we utilize the straight-through trick to stop the gradient of normalization item in Eq.~\eqref{eq:norm-deepspeed}, which  ensures that the gradient of routing scores remain valid:
 
\begin{equation}
    \frac{\partial L}{\partial a_{ij}} \equiv \frac{\partial L}{\sum_j g_{ij}\partial \frac{g_{ij}}{\sum_j g_{ij}}}\frac{\partial g_{ij}}{\partial a_{ij}}
    \label{eq:st-norm}
\end{equation}
 No modifications have been made to the forward stage. But at the backward stage, the gradient of the routing score $\frac{\partial L}{\partial g_{ij}}$ is calculated as $\frac{\partial L}{\sum_j g_{ij}\partial \frac{g_{ij}}{\sum_j g_{ij}}}$ rather than 0, where the normalization item $\sum_j g_{ij}$ is taken as a constant number without gradient.

\begin{figure*}[t]
  \centering

  \begin{subfigure}[b]{0.49\textwidth}
    \includegraphics[width=\textwidth]{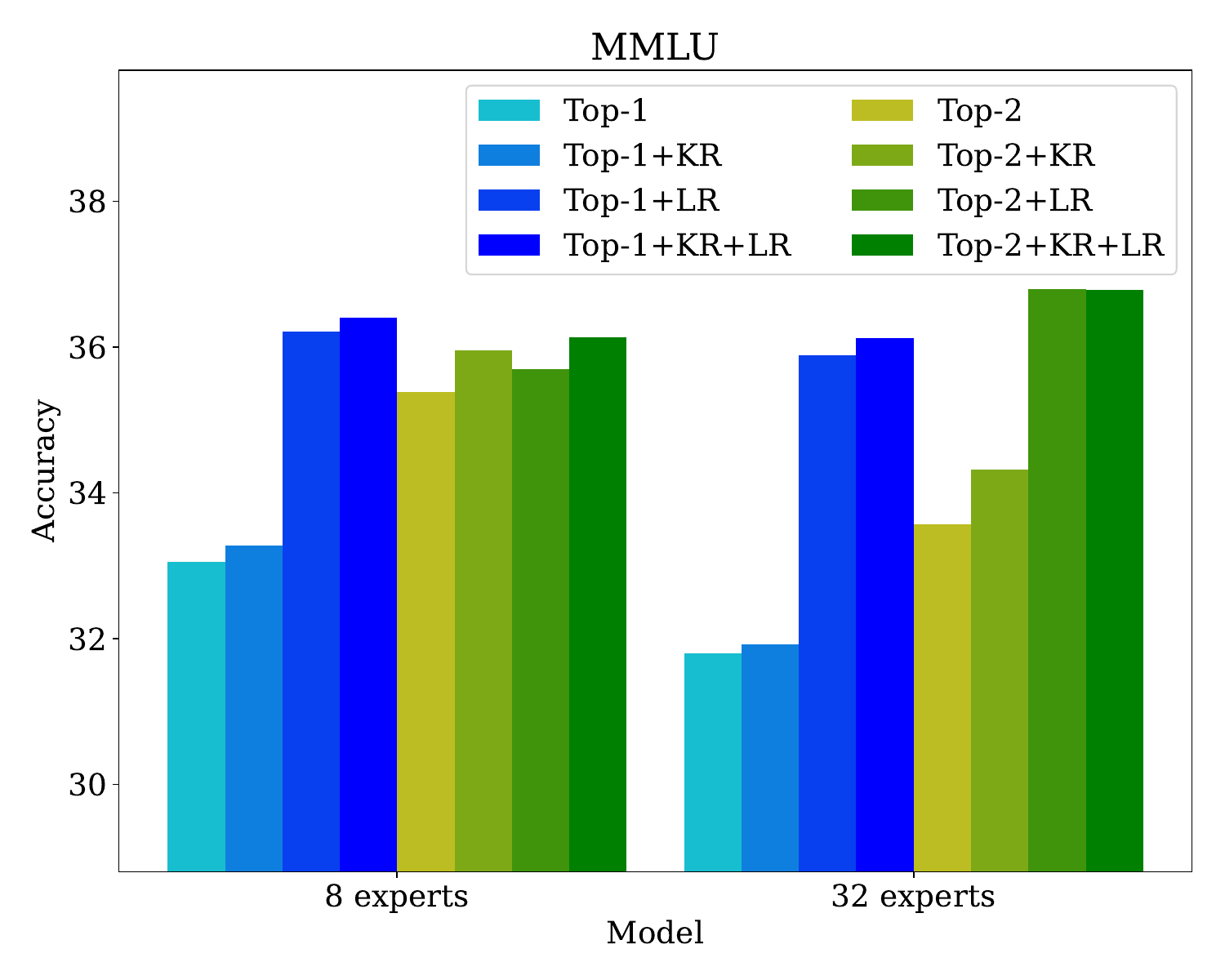}
    \caption{The performance of 8-experts and 32-experts MoEs on MMLU}
    \label{fig:scale1}
  \end{subfigure}
  \hfill
  \begin{subfigure}[b]{0.49\textwidth}
    \includegraphics[width=\textwidth]{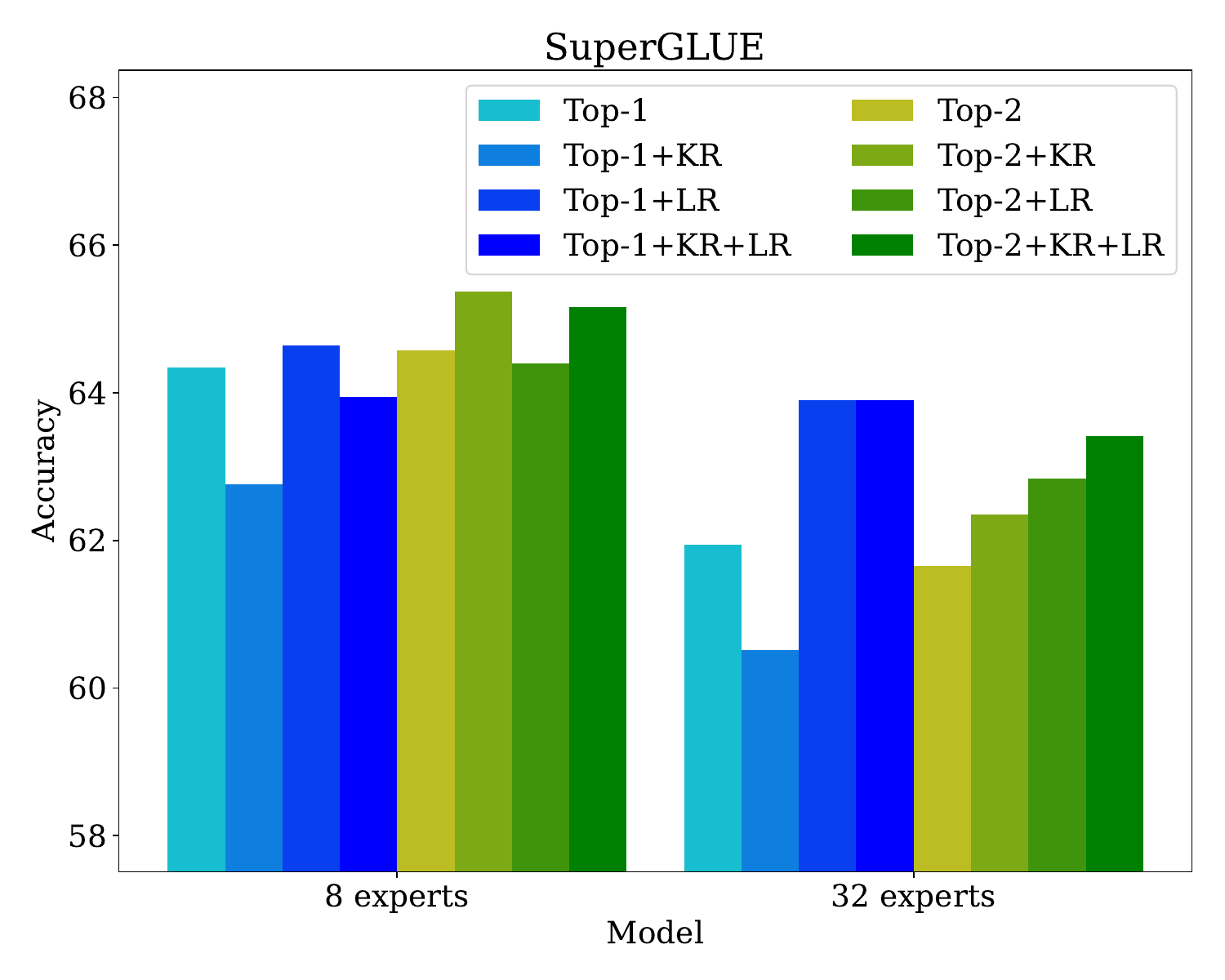}
    \caption{The performance of 8-experts and 32-experts MoEs on SuperGLUE}
    \label{fig:scale2}
  \end{subfigure}
  \begin{subfigure}[b]{0.49\textwidth}
    \includegraphics[width=\textwidth]{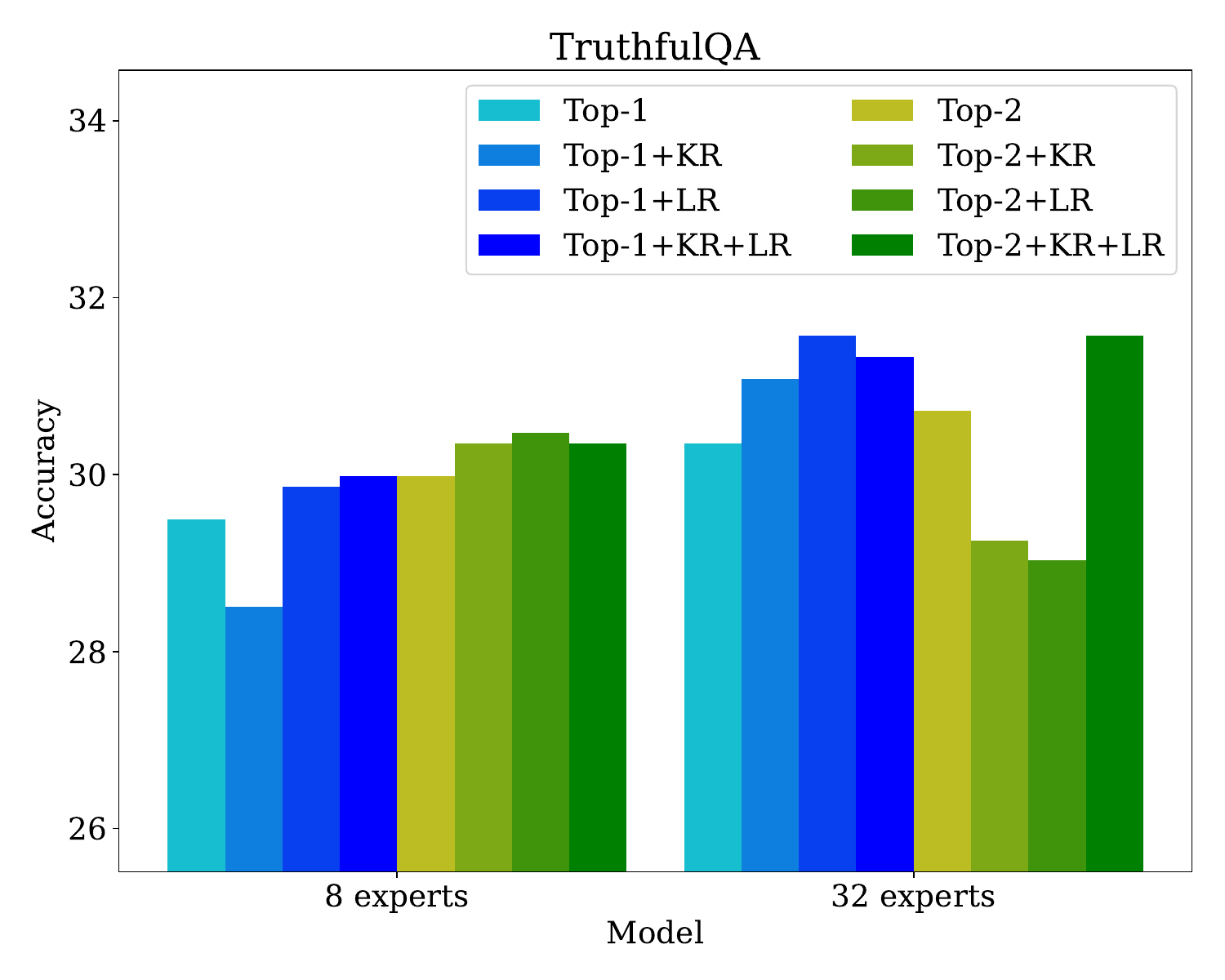}
    \caption{The performance of 8-experts and 32-experts MoEs on TruthfulQA}
    \label{fig:scale3}
  \end{subfigure}
  \hfill
  \begin{subfigure}[b]{0.49\textwidth}
    \includegraphics[width=\textwidth]{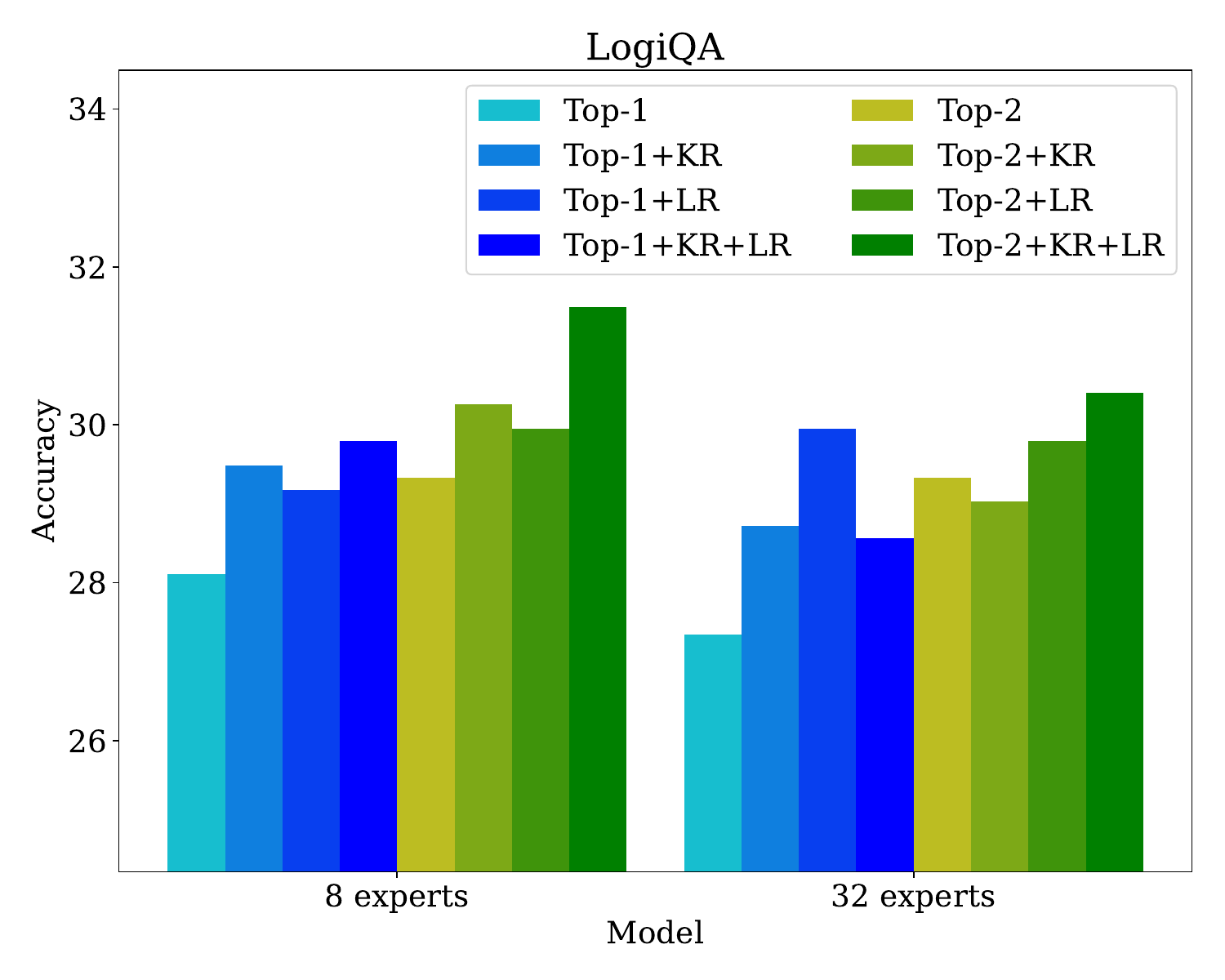}
    \caption{The performance of 8-experts and 32-experts MoEs on LogiQA}
    \label{fig:scale4}
  \end{subfigure}

  \caption{The performance of 8-experts and 32-experts MoEs on MMLU, SuperGLUE, TruthfulQA and LogiQA.}
  \label{fig:scale}
\end{figure*}

\section{Scaling to 32 Experts}\label{apd:scale-32}
In this experiment, we aimed to investigate the effectiveness of our methods when applied to a larger number of experts. We expanded the number of experts from 8 to 32. To reduce training costs, we only applied the Rectify-Routers (Intra-GPU Rectification and Fill-in Rectification) during evaluation. The results of this experiment are presented in Figure \ref{fig:scale}.

Interestingly, our findings indicate that increasing the number of experts from 8 to 32 does not necessarily result in improved model performance. In fact, in certain benchmarks, such as SuperGLUE, the performance of the model even declined. This observation aligns with previous research \citep{upcycling}, suggesting that increasing the number of experts can potentially be detrimental. One plausible explanation for this phenomenon is that a larger number of experts may lead to overfitting of the model. We believe that increasing the number of experts is helpful with enough training data. Notably, scaling from 8 to 32 experts only yielded notable benefits in the case of TruthfulQA.

Despite the lack of consistent improvement when increasing the number of experts, our methods (Intra-GPU Rectification and Fill-in Rectification) still demonstrated significant enhancements compared to the vanilla top-$k$ routing approach in the context of 32 experts. For instance, while the vanilla top-1 and top-2 routers with 32 experts underperformed those with 8 experts on MMLU, our methods (Top-2+FR+IR) enabled the 32-expert models to outperform their 8-expert counterparts.


\section{Analysis}
\subsection{Impact of Expert Distribution}\label{apd:expert-dist}
\begin{table*}[t]
    \centering
    \begin{tabular}{lllllll}
    \toprule
    \textbf{Router}                   & \textbf{Experts/GPU} & \textbf{MMLU}   & \textbf{SuperGLUE} & \textbf{TruthfulQA} & \textbf{LogiQA}  &\textbf{Avg}\\
\midrule
    \multirow{3}{*}{Top-1+IR} & 1               & 36.21 & 64.64    & 29.86     & 29.18   &39.97\\
                             & 2               & 36.17 & 64.38    & 30.23     & 29.03   &39.95\\
                             & 4               & 36.47 & 64.37    & 29.62     & 28.87   &39.83\\
\midrule
    \multirow{3}{*}{Top-2+IR} & 1               & 35.70 & 64.40    & 30.47     & 29.95   &40.13
\\
                             & 2               & 35.79 & 64.73    & 30.35     & 29.18   &40.01\\
                             & 4               & 35.79 & 65.38    & 30.35     & 29.18   &40.17
\\
    \bottomrule
    \end{tabular}%
    \caption{The performance of Intra-GPU Rectification evaluated under various settings of the number of experts per GPU.}
    \label{tab:topo}
\end{table*}

\begin{table*}[t]
    \centering
    \begin{tabular}{@{}llll@{}lll}
    \toprule
    \textbf{Model}   & \textbf{ST} & \textbf{MMLU} & \textbf{SuperGLUE}  &\textbf{TruthfulQA} &\textbf{LogiQA} &\textbf{Avg}\\ \midrule
    Top-1+FR & Yes    &      \textbf{34.66} & \textbf{63.97} & 28.51         & \textbf{29.18}  &\textbf{39.08}
\\
    Top-1+FR & No     &      33.96          &         62.75  & \textbf{29.25}& 28.57 &38.63\\
    \midrule
    Top-2    & Yes    &      35.39         &           64.58         & 29.98 & \textbf{29.33}   &\textbf{39.82}
\\
    Top-2    & No     &      \textbf{35.86}&           \textbf{64.73}& 29.98 & 28.26 &39.70\\ \bottomrule
    \end{tabular}%
    \caption{The performance of Top-1+FR and Top-2 router with and without straight-through trick. The second column (ST) denotes whether the straight-through trick is used.}
    \label{tab:st}
\end{table*}

Our Intra-GPU Rectification is a variant of the top-1 router, where tokens are assigned to the top-1 expert within GPU. When all experts are situated in the same GPU, the Intra-GPU Rectification essentially functions as the top-1 router. Therefore, the distribution of experts across GPUs can potentially influence the performance of the Intra-GPU Rectification. We conducted an investigation to explore this aspect and present the results in Table \ref{tab:topo}.

Interestingly, we found that increasing the number of experts per GPU did not yield significant improvements for either the top-1 router or the top-2 router. This suggests that the Intra-GPU Rectification demonstrates robustness to variations in the number of experts per GPU.

\subsection{Impact of Straight-through Trick}\label{apd:ablation-st}

In Appendix \ref{apd:grad-issue}, we propose a solution to address the gradient issue associated with the Fill-in Rectification by utilizing the straight-through trick. To evaluate the effectiveness of this technique, we conducted an experiment comparing the performance of the Fill-in Rectification with versus without the straight-through trick. The results of this comparison are presented in Table \ref{tab:st}.

Our findings indicate that the straight-through trick proves to be beneficial in improving the performance of the Fill-in Rectification (Top-1+FR). This suggests that the straight-through trick is necessary for the Fill-in Rectification to achieve optimal results. However, the application of the straight-through trick does not yield a significant improvement in the performance of the top-2 router. This can be attributed to the fact that the proportion of unsuccessful routing is relatively small (5\%) for the top-2 router, while it is considerably large (50\%) when employing the Fill-in Rectification.

\subsection{Impact of Load-Balance Loss}\label{apd:no-laux}
\begin{table*}[t]
    \centering
    \begin{tabular}{lllllll}
    \toprule
        \textbf{Model}  & \textbf{Aux-loss} & \textbf{MMLU} & \textbf{SuperGLUE}  & \textbf{TruthfulQA} & \textbf{LogiQA} &\textbf{Avg}\\ 
    \midrule
         Top-1& yes& 33.05 & \textbf{64.34} & \textbf{29.49} &28.11  &\textbf{38.74}\\
         Top-1& no& \textbf{33.88}& 61.60& 27.41&\textbf{29.95} &38.21
\\ \midrule
         Top-1+FR+IR& yes& \textbf{36.40}& 63.94& 29.98 & \textbf{29.80} &\textbf{40.03}
\\
         Top-1+FR+IR& no& 36.09& 64.05& \textbf{31.21} & 28.57 &39.98
\\ \bottomrule
    \end{tabular}
    \caption{The performance comparison of using vs. not using load-balance loss. Aux-loss represents whether load-balance loss is used.}
    \label{tab:laux}
\end{table*}
The Rectify-Routers proposed in this paper were designed to address the issues of dropped tokens and padding resulting from unbalanced routing. In our previous experiments, we utilized the load-balanced loss introduced by \citet{gshard} to enhance the balance of routing for all models, including those utilizing the Rectify-Routers. However, it is intriguing to investigate whether the Rectify-Routers remain effective in the absence of the load-balance loss. The results of this exploration are presented in Table \ref{tab:laux}.

Upon analyzing the results in Table \ref{tab:laux}, we observed a notable disparity in the performance of the vanilla top-1 router with and without the load-balance loss, particularly in the case of SuperGLUE and TruthfulQA. This discrepancy suggests that the load-balance loss plays a crucial role in improving the performance of the vanilla top-1 router. However, when considering our Rectify-Routers (Top-1+FR+IR), removing the load-balance loss does not result in a significant loss of performance. This finding indicates that our Rectify-Routers enhance the resilience of the top-1 router against the load-balance loss. Nevertheless, as a general trend, it is still preferable to employ a load-balance loss, even when utilizing the Rectify-Routers.

\section{Complete Results of Capacity Factor Variation}\label{app:complete}
In Section \ref{sec:capacity-var}, we have discussed the performance of MoE models across various capacity factor settings. However, it is worth noting that only the average accuracy are reported in Table \ref{tab:low-cf} and Table \ref{tab:high-cf}. For a comprehensive overview, we present the complete results in Table \ref{tab:low-cf-complete} and Table \ref{tab:high-cf-complete}, which encompass the evaluation outcomes across all benchmarks.

\begin{table*}[t]
    \centering
    \resizebox{1.0\textwidth}{!}{%
    \begin{tabular}{llllllllr}
    \toprule
    \textbf{Train Router} & \textbf{Test Router} & \textbf{Test CF} & \textbf{Test Speed} & \textbf{MMLU} & \textbf{SuperGLUE} & \textbf{TruthfulQA} & \textbf{LogiQA}  & \textbf{Avg}\\ \midrule
    \multirow{6}{*}{Top-1} & Top-1    & \multirow{2}{*}{1.0}  & 9.4k 
    & 33.05 &\textbf{64.34} & 29.49 & 28.11  &38.74\\
                           & Top-1+IR &                       & 9.2k & \textbf{36.21} & \textbf{64.64} & \textbf{29.86} & \textbf{29.18}  &\textbf{39.97}\\ \cmidrule(l){2-9} 
                           & Top-1    & \multirow{2}{*}{0.75}&  12.1k& 30.88 & 62.68 & 29.37 & 28.41  &37.83\\
                           & Top-1+IR &                       &  9.9k& \textbf{36.12} & \textbf{64.61} & \textbf{29.74} & \textbf{29.80}  &\textbf{40.06}\\ \cmidrule(l){2-9} 
                           & Top-1    & \multirow{2}{*}{0.5}&  16k& 26.43 & 59.71 & 26.68 & 26.57  &34.84\\
                           & Top-1+IR &                       &  10.6k& \textbf{36.32} & \textbf{65.23} & \textbf{29.98} & \textbf{30.10}  &\textbf{40.40}\\ \midrule
    \multirow{6}{*}{Top-2} & Top-2    & \multirow{2}{*}{2.0}  &  6.2k & 35.39 &  \textbf{64.58} & 29.98 & 29.33   &39.82
    \\
                           & Top-2+IR &                       &  6.0k & \textbf{35.70} & 64.40 & \textbf{30.47} & \textbf{29.95}  &\textbf{40.13}\\ \cmidrule(l){2-9} 
                           & Top-2    & \multirow{2}{*}{1.5}  &  7.4k& 35.22 & \textbf{65.00} & \textbf{30.47} & 27.34  &39.50\\
                           & Top-2+IR &                       &  6.6k& \textbf{35.71}& 64.47 & 29.98 & \textbf{28.26}  &\textbf{39.60}\\ \cmidrule(l){2-9} 
                           & Top-2    & \multirow{2}{*}{1.0}  &  8.9k &  33.28 &  62.76 & 28.51 &  29.49  &38.51
    \\
                           & Top-2+IR &                       &  7.3k&  \textbf{35.93}&  \textbf{64.40}& \textbf{29.62}&  \textbf{30.10} &\textbf{40.01}\\ \midrule
    \end{tabular}%
    }
    \caption{Performance of top-$k$ routers and their variants with low capacity factors ($<=k$). The difference between this table and Table \ref{tab:low-cf} is that the evaluation results on all benchmarks are reported in this table, but only the average accuracy is reported in Table \ref{tab:low-cf}}
    \label{tab:low-cf-complete}
\end{table*}

\begin{table*}[t]
    \centering
    \resizebox{0.9\textwidth}{!}{%
    \begin{tabular}{llllllllr}
    \toprule
    \textbf{Train Router} & \textbf{Test Router} & \textbf{Test CF}  &\textbf{Test Speed}& \textbf{MMLU} & \textbf{SuperGLUE} & \textbf{TruthfulQA} & \textbf{LogiQA}  & \textbf{Avg}\\ \midrule
    \multirow{6}{*}{Top-1} & Top-1     & \multirow{2}{*}{1.0}   &9.4k & 33.05 &\textbf{64.34} & \textbf{29.49} & 28.11  &\textbf{38.74}\\
                           & Top-1+FR&                        &8.9k& \textbf{33.28} & 62.76 & 28.51 & \textbf{29.49}  &38.51
    \\\cmidrule(l){2-9} 
                           & Top-1     & \multirow{2}{*}{1.25}  &8.6k& 34.51 & 62.94 & 29.13 & \textbf{31.79}  &39.59\\
                           & Top-1+FR&                        &8.1k& \textbf{35.01} & \textbf{65.22} &\textbf{30.23} & 29.95  &\textbf{40.10}\\\cmidrule(l){2-9} 
                           & Top-1     & \multirow{2}{*}{1.5}   &7.9k& \textbf{36.40} & 64.21 & 28.88 & 29.95   &39.86
    \\
                           & Top-1+FR&                        &7.3k& 36.10 & \textbf{64.89} & \textbf{30.23} & \textbf{30.10}  &\textbf{40.33}\\ \midrule
    \multirow{6}{*}{Top-2} & Top-2     & \multirow{2}{*}{2.0}   &6.2k & 35.39 & 64.58 & 29.98 & 29.33   &39.82
    \\
                           & Top-2+FR&                        &5.8k & \textbf{35.96}& \textbf{65.37}& \textbf{30.35}&	\textbf{30.26} &\textbf{40.48}\\ \cmidrule(l){2-9} 
                           & Top-2     & \multirow{2}{*}{2.5}   &5.4k& 35.96 & 64.53 & 30.23 & 28.87  &39.89\\
                           & Top-2+FR&                        &5.1k& 36.00 & \textbf{64.92}& \textbf{30.59}& \textbf{30.56} & \textbf{40.51}\\ \cmidrule(l){2-9} 
                           & Top-2     & \multirow{2}{*}{3.0}   &4.9k& 35.72 & 64.65 & \textbf{30.59}& 29.18  &40.03\\
                           & Top-2+FR&                        &4.5k& \textbf{35.98}&  \textbf{65.31}& 30.23 & \textbf{30.26} &\textbf{40.44}\\
    \bottomrule
    \end{tabular}%
    }
    \caption{Performance of top-$k$ routers and their variants with high capacity factors ($>=k$). The difference between this table and Table \ref{tab:high-cf} is that the evaluation results on all benchmarks are reported in this table, but only the average accuracy is reported in Table \ref{tab:high-cf}}
    \label{tab:high-cf-complete}
\end{table*}




\end{document}